\documentclass[10pt,twocolumn,letterpaper]{article}

\usepackage{cvpr}      
\usepackage[dvipsnames]{xcolor}

\definecolor{cvprblue}{rgb}{0.21,0.49,0.74}
\usepackage[pagebackref,breaklinks,colorlinks,citecolor=cvprblue]{hyperref}

\def\paperID{9034} 
\def\confName{CVPR}
\def\confYear{2024}

\title{Rethinking Multi-domain Generalization with A General Learning Objective}

\author{Zhaorui Tan$^{1,2}$, Xi Yang$^*$$^{1}$, Kaizhu Huang\thanks{Corresponding authors}~~$^{3}$\\
$^{1}$Xi’an Jiaotong-Liverpool University, $^{2}$University of Liverpool, $^{3}$Duke Kunshan University\\
{\tt\small Zhaorui.Tan21@student.xjtlu.edu.cn, Xi.Yang01@xjtlu.edu.cn, kaizhu.huang@dukekunshan.edu.cn}
}

\usepackage{amsmath,amsfonts,bm}

\def\Figref#1{Figure~\ref{#1}}

\def\Secref#1{Section~\ref{#1}}

\def\eqref#1{equation~\ref{#1}}

\def\1{\bm{1}}

\def\rmX{{\mathbf{X}}}
\def\rmY{{\mathbf{Y}}}
\def\rmZ{{\mathbf{Z}}}

\def\vx{{\bm{x}}}
\def\vy{{\bm{y}}}

\DeclareMathAlphabet{\mathsfit}{\encodingdefault}{\sfdefault}{m}{sl}
\SetMathAlphabet{\mathsfit}{bold}{\encodingdefault}{\sfdefault}{bx}{n}

\newcommand{\KL}{D_{\mathrm{KL}}}

\usepackage{enumitem}
\setlist[itemize]{align=parleft,left=0pt..1em,itemsep=0pt,parsep=0pt,topsep=0pt}
\usepackage{algorithm}
\usepackage{algorithmic}
\usepackage{amsmath}
\usepackage{amssymb}
\usepackage{amsthm}
\usepackage{amsfonts}
\usepackage{color}
\newcommand\blue{\textcolor{blue}}
\usepackage{hyperref}
\usepackage{url}
\usepackage{graphicx}

\newtheorem{definition}{Definition}

\newtheorem{remark}{Remark}
\newcommand{\indep}{\perp\!\!\!\perp}
\usepackage{booktabs}
\usepackage{color}
\usepackage{multirow}

\usepackage{wrapfig}
\usepackage{colortbl}
\definecolor{mygray}{gray}{.85}

\usepackage[capitalize]{cleveref}
\crefname{section}{Sec.}{Secs.}
\crefname{table}{Tab.}{Tabs.}

\begin{document}
\maketitle

\begin{abstract}
Multi-domain generalization (mDG) is universally aimed to minimize the discrepancy between training and testing distributions to enhance marginal-to-label distribution mapping. However, existing mDG literature lacks a general learning objective paradigm and often imposes constraints on static target marginal distributions.
In this paper, we propose to leverage a $\rmY$-mapping 
to relax the constraint. We rethink the learning objective for mDG and design a new \textbf{general learning objective} to interpret and analyze most existing mDG wisdom. This general objective is bifurcated into two synergistic amis: learning domain-independent conditional features and maximizing a posterior. 
Explorations also extend to two effective regularization terms that incorporate prior information and suppress invalid causality, alleviating the issues that come with relaxed constraints. We theoretically contribute an upper bound for the domain alignment of domain-independent conditional features, disclosing that many previous mDG endeavors actually \textbf{optimize partially the objective} and thus lead to limited performance.
As such, our study distills a general learning objective into four practical components, 
providing a general, robust, and flexible mechanism to handle complex domain shifts. Extensive empirical results indicate that the proposed objective with $\rmY$-mapping leads to substantially better mDG performance in various downstream tasks, including regression, segmentation, and classification. Code is available at \url{https://github.com/zhaorui-tan/GMDG/tree/main}.
\end{abstract}

\section{Introduction}

Domain shift, which breaks the independent and identical distributed (\textit{i.i.d.}) assumption amid training and test distributions~\citep{wang2022generalizing}, poses a common yet challenging problem in real-world scenarios.
Multi-domain generalization (mDG)~\citep{blanchard2011generalizing}) is garnering increasing attention owing to its promising capacity to utilize multiple distinct but related source domains for model optimization, ultimately intending to generalize well to unseen domains.
Intrinsically, the primary objective for mDG is the maximization of the joint distribution between observations $\rmX$ and targets $\rmY$ across all domains $\mathcal{D}$:
\begin{equation}
    \begin{split}
            \label{eq:obj1}
    \max P({\rmX,\rmY \mid \mathcal{D}} ) 
    = & P({\rmY \mid \mathcal{D}}) P ({\rmX \mid \rmY,\mathcal{D}}) \\
    =& P({\rmX \mid \mathcal{D}}) P ({\rmY \mid \rmX,\mathcal{D}}).
    \end{split}
\end{equation}
A prevalent approach initiates by maximizing the marginal distribution $P(\rmX|\mathcal{D})$ before presuming an invariant $P(\rmY|\rmX)=P(\rmY|\rmX, \mathcal{D})$ across domains~\citep{zhou2022domain},
anchored on an assumption that $P(\rmY|\mathcal{D})$ remains consistency across $\mathcal{D}$. 

\begin{table}[t]
\centering
\resizebox{0.9\linewidth}{!}{%
\begin{tabular}{l|l}
\toprule
 & \multicolumn{1}{l}{\bf Aim1: Learning domain invariance}  \\ 
Others & None  \\
DANN & $\min_{\phi} H(  P (\phi(\rmX) \mid \mathcal{D}))$  \\
CDANN, CIDG, MDA & $\min_{\phi} H(  P ({\phi(\rmX), \rmY} \mid \mathcal{D}))$  \\
\rowcolor{mygray} \textbf{Ours: GAim1} & {$\min_{\phi, \psi} H( P ({\phi(\rmX), \psi(\rmY)} \mid \mathcal{D}))$} \\
\bottomrule 
\toprule
 &   \multicolumn{1}{l}{\bf Reg1: Integrating prior} \\ 
Others & None \\
MIRO, SIMPLE & {${\min_{\phi}\KL(P({\phi(\rmX), \rmY})\Vert \mathcal{O})}$} \\
\rowcolor{mygray} \textbf{Ours: GReg1}  & {${\min_{\phi,\psi}\KL(P({\phi(\rmX), \psi{(\rmY)}})\Vert \mathcal{O})}$} \\
\bottomrule 
\toprule
 & \multicolumn{1}{l}{ \bf  Aim2: Maximizing A Posterior (MAP)}   \\ 
Others &{$\min_{\phi}H(P ({\rmY ,  \phi(\rmX)}))$}  \\
\rowcolor{mygray} \textbf{Ours: GAim2} &{$\min_{\phi,\psi}H(P ({\rmY ,   \phi(\rmX)})) + H(P ({\rmY ,   \psi(\rmY)}))$}   \\
\bottomrule
\toprule
   & \multicolumn{1}{l}{ \bf  Reg2: Suppressing invalid causality}\\
Others  & None \\
CORAL   & {$\min_{\phi}-H( P ({\phi(\rmX) \mid  \mathcal{D} })) + H(P({\phi(\rmX)}))$} \\
MDA,RobutsNet  & {$\min_{\phi} -  H ( P({\phi(\rmX)\mid \rmY}))  +  H(P (\phi(\rmX))) $} \\
\rowcolor{mygray} \textbf{Ours: GReg2}  & {${\min_{\phi,\psi} - H(P(\phi(\rmX)  \mid  \psi(\rmY))) +H(P(\phi(\rmX)))}$}  \\
\bottomrule
\end{tabular}%
}
\vspace{-5pt}
\caption{A summary of objectives of  ERM~\citep{gulrajani2020search}, 
DANN~\citep{ganin2016domain},
CORAL~\citep{sun2016deep},
CDANN~\cite{li2018deep}, 
CIDG~\citep{li2018domain}, MDA~\citep{hu2020domain}, MIRO~\citep{cha2022miro}, SIMPLE~\citep{li2022simple},
RobustNet~\citep{choi2021robustnet},
VA-DepthNet~\cite{liu2023va}
and Ours. 
All constants are omitted here. `Others' denotes no other specified methods.  For more details, see Supplementary Material~\ref{app:our_obj}.
}
\label{tab:summary}
\vspace{-8pt}
\end{table}

\textit{Is $P(\rmY|\mathcal{D})$ truly static across domains?} In other words, does $\rmY$ truly lack domain-dependent features?
In classification tasks, typically, the influence of $\mathcal{D}$ on $\rmY$ is substantially marginal. However, this assumption is not universally applicable, particularly in tasks such as regression or segmentation. Consequently, MDA~\citep{hu2020domain} relaxes the assumption of stable $P(\rmY|\mathcal{D})$ by providing an average class discrepancy, allowing both $P(\rmX|\rmY, \mathcal{D})$ and $P(\rmY|\mathcal{D})$ vary across $\mathcal{D}$. However, MDA has to conduct class-specific sample selection under domains for obtaining $P(\rmX|\rmY,\mathcal{D})$, which constrains its objective's universality and struggles with tasks beyond basic classification, especially where $\rmY$ is not discrete.

To better tackle the $\mathcal{D}$-dependent variations in both $\rmX$ and $\rmY$ for border tasks beside classification, we introduce two learnable mappings, $\phi$ and $\psi$, that project $\rmX$ and $\rmY$ into the \textbf{same} latent Reproducing Kernel Hilbert Space (RKHS), assumed to extract $\mathcal{D}$-independent features from $\rmX,\rmY$. Incorporating these, Eq.~\ref{eq:obj1} can be changed as
\begin{align}
    \label{eq:obj2}
    \max\nolimits_{\phi,\psi} P ( \phi(\rmX), \psi(\rmY)) 
    , \; \text{ s.t., } \phi(\rmX), \psi(\rmY) \indep \mathcal{D}.
\end{align}
Built upon the optimization of Eq.~\ref{eq:obj2}, we further identify two additional issues that warrant consideration. 1). The synergy of \textit{integrating prior information} and domain-invariant feature learning plays a crucial role. pre-trained (oracle) models can be used as priors~\citep{cha2022miro, li2022simple}  to regulate feature learning.   
{2). Issues regarding \textit{invalid causality predicament} within the $P(\rmY|\mathcal{D})$ static assumption relaxation during learning the invariance come to light.} This is aligned with the causality premise $\phi (\rmX) \to \psi (\rmY)$ to maximize $P ({\phi(\rmX), \psi(\rmY) \mid \mathcal{D}})$. Efforts must be made to suppress invalid causality $ \psi(\rmY)\to\phi(\rmX)$ during invariant-feature learning (Refer to Eq.~\ref{eq:GAim1_GReg2} for derivation).

Considering these findings, the general objective for mDG, which copes with the above issues and effectively relaxes the static target distribution assumption, is crucial. 
To be specific, it should consist of \textbf{four key parts}:  
\textbf{Aim1}- Learning domain-invariant representations and \textbf{Aim2}- Maximizing the posterior; with two regularization \textbf{Reg1}- Integrating prior information and the \textbf{Reg2}- Suppression of invalid causality. In essence, the objective should certify invariant representations of $\rmX,\rmY$ across domains while preserving the prediction relationship in $\rmX \! \to \!\rmY$. As a notable contribution, we redesign the conventional mDG paradigm and uniformly simplify most previous works' empirical objectives, as summarized in Table~\ref{tab:summary} while Notations are shown in Table~\ref{tab:notations_of_sym}.   


\begin{table}[t]
    \centering
    \resizebox{0.9\linewidth}{!}{%
            \begin{tabular}{rp{8.5cm}}
            \toprule
            \multicolumn{1}{c}{\bf Symbols }  &\multicolumn{1}{c}{\bf Descriptions}
            \\ \midrule
            $d_n \in \mathcal{D}, n \leq N$ ; $d' \in \mathcal{D}$:           & The $n$-th observed domains in all domains;  Unseen domains in  all domains.  \\
            $\rmX, \rmY$; $\rmX_n, \rmY_n$; $\rmX', \rmY'$.           & All observations and targets; Observations and targets in $d_{n}$; Observations and targets in $d'$.  \\
            $P(x)$: &  Distributions where $x$ corresponds to the random variables. \\
            $\phi, \psi$: &  Learnable transformations that codify $\rmX, \rmY$ into the same latent RKHS. \\
            $\phi(\rmX), \psi(\rmY)$:  & Mapped $\rmX, \rmY$. Within the RKHS realm, $\phi(\rmX), \psi(\rmY)$ follow Multivariate Gaussian Distributions. \\
            $\mathcal{O}$; $R(\cdot)$; $\sigma_{\cdot,\cdot}$: & Prior knowledge (oracle model); Empirical risks; Covariance between two variables.\\
            $\mathcal{C}: \phi(\rmX), \psi(\rmY) \to \rmY$: & Predictor that predicts $\rmY$ from $\phi(\rmX), \psi(\rmY) $. \\
            $\KL(\cdot \Vert \cdot); H(\cdot); H_c(\cdot,\cdot)$: &  KL divergence; Entropy; Cross-entropy. \\
            \bottomrule
            \end{tabular}%
            }
            \vspace{-5pt}
            \caption{A summary of notations.}
            \label{tab:notations_of_sym}
\vspace{-8pt}
\end{table}

Most current mDG studies only focus on classification. SOTA methods such as MIRO~\citep{cha2022miro} and SIMPLE~\citep{li2022simple} propose learning similar features by ``oracle" models as a substitute for learning domain-invariant representations for mDG. Worth mentioning, we counter MIRO's argument by confirming the persisting necessity of domain-invariant features, even under prior distribution, by theoretically deviating from minimizing the Generalized Jensen-Shannon Divergence (GJSD). MDA~\citep{hu2020domain} pioneered the relaxation of the $P(\rmY|\mathcal{D})$ static assumption,
yet without explicitly introducing a $\rmY$-mapping function, and overlooked the emergence of invalid causality that arises upon the relaxation. 
Beyond classification,
RobustNet~\citep{choi2021robustnet} and VA-DepthNet~\cite{liu2023va} explore their methods on mDG settings in segmentation and regression but propose no explicit objective for mDG.
Importantly, our theoretical analysis and empirical findings suggest that mere aggregation of all the aforementioned objectives fails to yield a comprehensive general objective for mDG. 
For instance, term $\!-H( P ({\phi(\rmX|\mathcal{D} )}))$, coupled with prior knowledge utilization, could inadvertently precipitate performance degradation.

\begin{figure}[!t]
    \centering
    \includegraphics[width=0.8\linewidth]{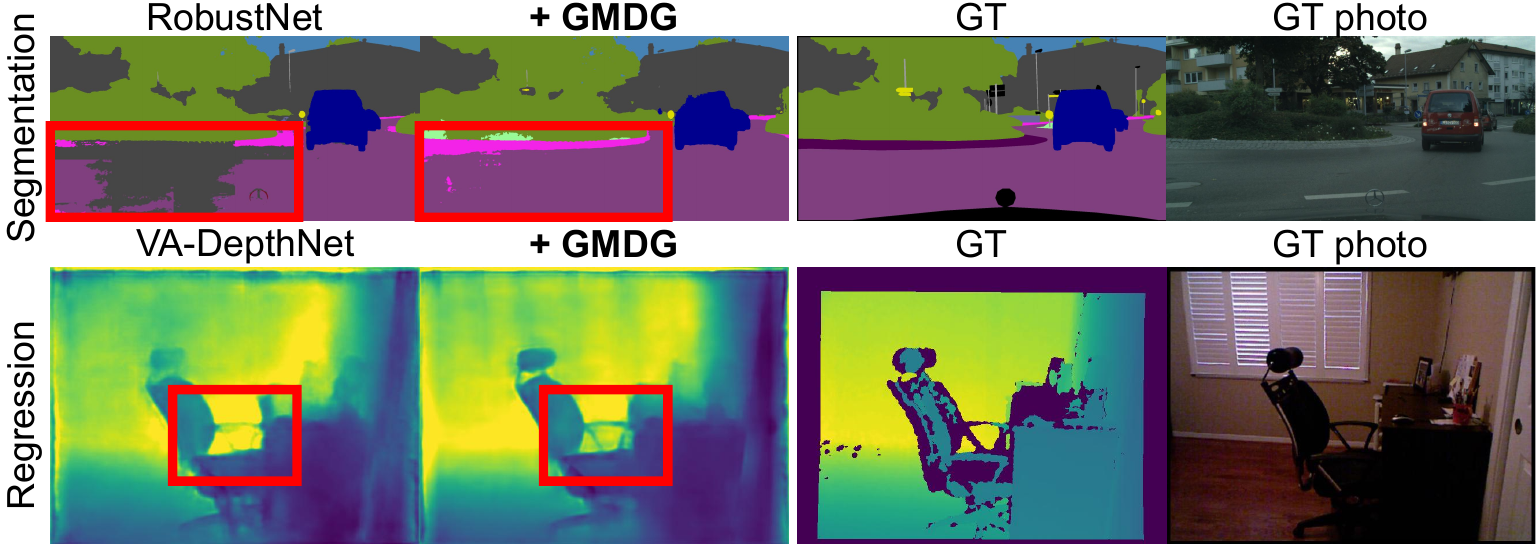}
    \vspace{-5pt}
    \caption{Segmentation and regression results of baselines and +GMDG on samples in unseen domains. 
    }
    \label{fig:seg_reg_vis}   
\vspace{-8pt}
\end{figure}

In this paper, we introduce the General Multi-Domain Generalization Objective (GMDG) to overcome current limitations in current methods,
relaxing the static assumption of $P(\rmY|\mathcal{D})$ (overall formulation is shown in \Secref{sec:dg}). Meanwhile, we propose an actionable solution to the invalidated causality through the minimization of the Conditional Feature Shift (CFS). Our main contributions can be summarized as follows:
\begin{itemize}
     \item We theoretically prove that domain generalization can be improved through the minimization of Generalized Jensen-Shannon Divergence (GJSD), with the incorporation of prior knowledge, leading to the derivation of an alignment upper bound (PUB) (\Secref{sec:dg}). 
     \item We analyze existing approaches, demonstrating their incomplete optimization against the GMDG and identifying unexpected terms they inadvertently introduce (\Secref{sec:conn}).
     \item Our approach is the first try that is designed as compatible with existing mDG frameworks and exhibits performance improvements in a suite of tasks, including regression, segmentation, and classification, as confirmed by our comprehensive experiments (\Secref{sec:exp}). 
\end{itemize}
Notably, our results that only used one pre-trained model as prior in classification tasks exceed the SOTA SIMPLE++, which employs 283 pre-trained models as an ensemble oracle, while yielding consistent improvement in regression and segmentation, as shown in Figure~\ref{fig:seg_reg_vis}.
This further suggests the superiority of GMDG.

\section{Related work}

\textbf{Multi-domain generalization.} 
Most current mDG methods focus only on classification tasks.
To learn better $D$-independent representations for mDG, DANN~\citep{ganin2016domain} minimizes feature divergences between the source domains. 
CDANN~\cite{li2018deep}, CIDG~\citep{li2018domain}, and MDA~\citep{hu2020domain} additionally take conditions into consideration and aim to learn conditionally invariant features across domains.
\citep{bui2021exploiting,chattopadhyay2020learning,cha2022miro,li2022simple} point out that learning invariant representation to source domains is insufficient for mDG. Thus, MIRO~\citep{cha2022miro} and SIMPLE~\citep{cha2022miro} adopt pre-trained models as an oracle for seeking better general representations across various domains, including unseen target domains.
Meanwhile, RobustNet~\cite{choi2021robustnet} constrains conditional covariance shifts and conducts mDG segmentation, and little exploration focuses on mDG regression, though many other works pay attention to single domain generalization~\cite{peng2022semantic,kang2022style,lee2022wildnet,liu2023va}. Our study shows that their objectives optimize partially GMDG, leading to sub-optimal results.


\textbf{Multi-domain generalization assumptions.} 
In the literature, different assumptions are proposed to simplify the task as described by the original objective in Eq.~\ref{eq:obj1}. One assumption is that the 
$P(\rmY|\rmX,\mathcal{D})$ is stable while only marginal $P(\rmX|\mathcal{D})$ changes across domains~\citep{shimodaira2000improving,zhang2015multi}. 
\cite{zhang2013domain}  point out that $\rmX$ is usually caused by $\rmY$ thus $P(\rmY|\mathcal{D})$ changes while $P(\rmX|\rmY,\mathcal{D})$ is sable or $P(\rmX|\rmY,\mathcal{D})$ changes but $P(\rmY|\mathcal{D})$ stays stable, or a combination of both. Thus, MDA~\citep{hu2020domain} allows both $P(\rmY|\rmX, \mathcal{D})$ and $P(\rmX| \mathcal{D})$ change across domains 
but needs selecting samples of each class for the calculation. Moreover,  it considers no prior. 
This paper further relaxes these assumptions by extracting domain-invariant features in $\rmX, \rmY$. 

\textbf{Using pre-trained models as an oracle.} Previous methods such as MIRO~\citep{cha2022miro} have employed pre-trained models as the oracle  to regularize $\phi$. SIMPLE~\citep{li2022simple} employs at most $283$ pre-trained models as an ensemble and adaptively composes the most suitable oracle model. 
RobustNet~\cite{choi2021robustnet} and VA-DepthNet~\cite{liu2023va}, only use pre-trained models as initialization rather than additional supervision.

\section{A general multi-Domain generalization objective} 
\label{sec:dg}

 
General Multi-Domain Generalization Objective (GMDG) essentially comprises a weighted combination of \textbf{Four} terms, 
each term designated by an alias:
\begin{equation}
    \resizebox{0.9\hsize}{!}{$\begin{split}
    \label{eq:dg_obj_final}
        \min\nolimits_{\phi, \psi} & \underbrace{v_{A1}  H( P ({\phi(\rmX), \psi(\rmY)} \mid \mathcal{D}))}_{\text{\textbf{GAim1}}} \\  
    & + 
   \underbrace{v_{A2} [H(P ({\psi(\rmY) ,   \phi(\rmX)})) + H(P ({\rmY ,  \psi(\rmY)}))]}_{\text{\textbf{GAim2}}}  \\
   &  + \underbrace{v_{R1} \KL(P({\phi(\rmX), \psi(\rmY)})\Vert \mathcal{O})}_{\text{\textbf{GReg1}}}   \\
   & \underbrace{ - v_{R2} H(P(\phi(\rmX)  \mid  \psi(\rmY))) +H(P(\phi(\rmX)))}_{\text{\textbf{GReg2}}}. 
    \end{split}$}
\end{equation}
Theoretically, we justify that \textbf{GAim1} and \textbf{GReg1} can be effectively revised by minimizing the Generalized Jensen-Shannon Divergence (GJSD) with prior knowledge between visible domains for optimization. Meanwhile,  we derive an upper bound termed as an alignment Upper Bound with Prior of mDG (PUB).
Importantly, we demonstrate using \textbf{GReg2} not only cope with the invalid causality brought by $P(\rmY|\mathcal{D})$ static assumption relaxation. 
Regarding \textbf{GReg2}, 
it can be simplified by minimizing the Conditional Feature Shift (CFS), \textit{i.e.,} the shift between unconditional and conditional features, which can be calculated by $\psi$. 
More theoretical details are provided as follows.

\subsection{Theoretical Details}
\textbf{Learning of $\mathcal{D}$-independent conditional features under prior.} The generalization alignment upper bound (PUB), a novel GJSD variational upper bound that is tied to domain generalization alignment, is derived based on the generalized Jensen-Shannon divergence (GJSD)~\citep{GJSD}.
\begin{definition}[GJSD] Given $J$ distributions, $\{P ({\rmZ_j})\}_{j=1}^{J}$ and a corresponding probability weight vector $w$, $GJSD_{w}(\{P ({\rmZ_j})\}_{j=1}^{J})$ is defined as:
\begin{equation}
    \begin{split}
          & \sum\nolimits_{j=1}^{J} w_j \KL(P ({\rmZ_j}) \Vert \sum\nolimits_{j=1}^{J} w_j P ({\rmZ_j})) \\
      \equiv  & H(\sum\nolimits_{j=1}^{J} w_j P( {\rmZ_j})) - \sum\nolimits_{j=1}^{J} w_j H (P ({\rmZ_j})). 
    \end{split}
\end{equation}
\end{definition}
Our method addresses the standard scenario in which the weights are evenly distributed across domains: $w_1=...=w_N = 1/N$. 
To achieve $ \phi(\rmX), \psi(\rmY)\indep\mathcal{D}$,
minimizing domain gap between $P ({\phi(\rmX_n), \psi(\rmY_n)})$ can be converted to minimizing GJSD across all domains:
\begin{equation}
    \begin{split}
        \label{eq:bound1}
    &\min\nolimits_{\phi, \psi} GJSD(\{P ({\phi(\rmX_n), \psi(\rmY_n)})\}_{n=1}^{N}) \\ 
     \equiv &\min\nolimits_{\phi, \psi}  H( P ({\phi(\rmX), \psi(\rmY)} \mid \mathcal{D}))  \\
     & - \mathbb{E} [H (P ({\phi(\rmX_n), \psi(\rmY_n)}))]. 
    \end{split}
\end{equation}

We further involve a prior knowledge distribution $\mathcal{O}$ under the consideration of a variational density model class $\mathcal{Q}$.
Drawing upon~\citep{cho2022cooperative},
we have a variational upper bound:
\begin{equation}
    \begin{split}
        \label{eq:bound2}
     &GJSD(\{P ({\phi(\rmX_n), \psi(\rmY_n)})\}_{n=1}^{N}) \\  \leq 
     &H_c(\mathbb{E} [P ({\phi(\rmX), \psi(\rmY) ]\mid \mathcal{D}}),\mathcal{O})-a, 
    \end{split}
\end{equation}
where $a \triangleq \sum\nolimits_{n=1}^N  H (P ({\phi(\rmX_n), \psi(\rmY_n)}))$ is constant \textit{w.r.t} $\phi, \psi$, hence ignored during optimization.
The novel PUB is derived from Eq.~\ref{eq:bound2}, is:
\begin{equation}
    \begin{split}
        \label{eq:PUB}
&  \min\nolimits_{\phi, \psi}PUB(\{P ({\phi(\rmX_n), \psi(\rmY_n)})\}_{n=1}^{N})
\\
\triangleq & \min\nolimits_{\phi, \psi}{ P ({\phi(\rmX), \psi(\rmY)} \mid \mathcal{D}))} \\
& + {\KL( P ({\phi(\rmX), \psi(\rmY)}) \Vert\mathcal{O}))}  -{a}.  
    \end{split}
\end{equation}

Minimizing PUB is the proposed objective for \textbf{GAim1} and \textbf{GReg1}.
This implies that methods like MIRO, solely minimizing GReg1, might result in substantial suboptimality, leaving the domain gap unresolved. 
We discuss two situations of $\mathcal{O}$ in \Secref{sec:conn}.


\textbf{Suppressing invalid causality.} 
{The relaxation of $P(\rmY|\mathcal{D})$ static assumption may lead to unexpected causality while learning the invariance.}
\textbf{GAim1} is reformed as:
\begin{equation}
    \begin{split}
        \label{eq:GAim1_GReg2}
    \text{\textbf{GAim1}} 
    = & H(P (\phi(\rmX ) \! \mid \! \mathcal{D}))\!\! +\!\! H(P(\psi(\rmY)\! \mid \! \phi(\rmX), \mathcal{D})) \\ 
    = & H(P (\psi(\rmY) \! \mid \! \mathcal{D})) \!\!+ \!\! H(P(\phi(\rmX)\! \mid \!\psi(\rmY),  \mathcal{D})),
    \end{split}
\end{equation}
{where minimizing \textbf{GAim1} with relaxation of $P(\rmY|\mathcal{D})$ static assumption may lead to $\psi(\rmY) \! \to \!\phi(\rmX)$ since the term $H( P ( \psi(\rmY) | \phi(\rmX), \mathcal{D}))$ and $\phi (\rmX) \!\to \! \psi(\rmY)$ since the term $H(P(\phi(\rmX) | \psi(\rmY),  \mathcal{D}))$.}
\begin{figure}
\centering
\includegraphics[width=0.8\linewidth]{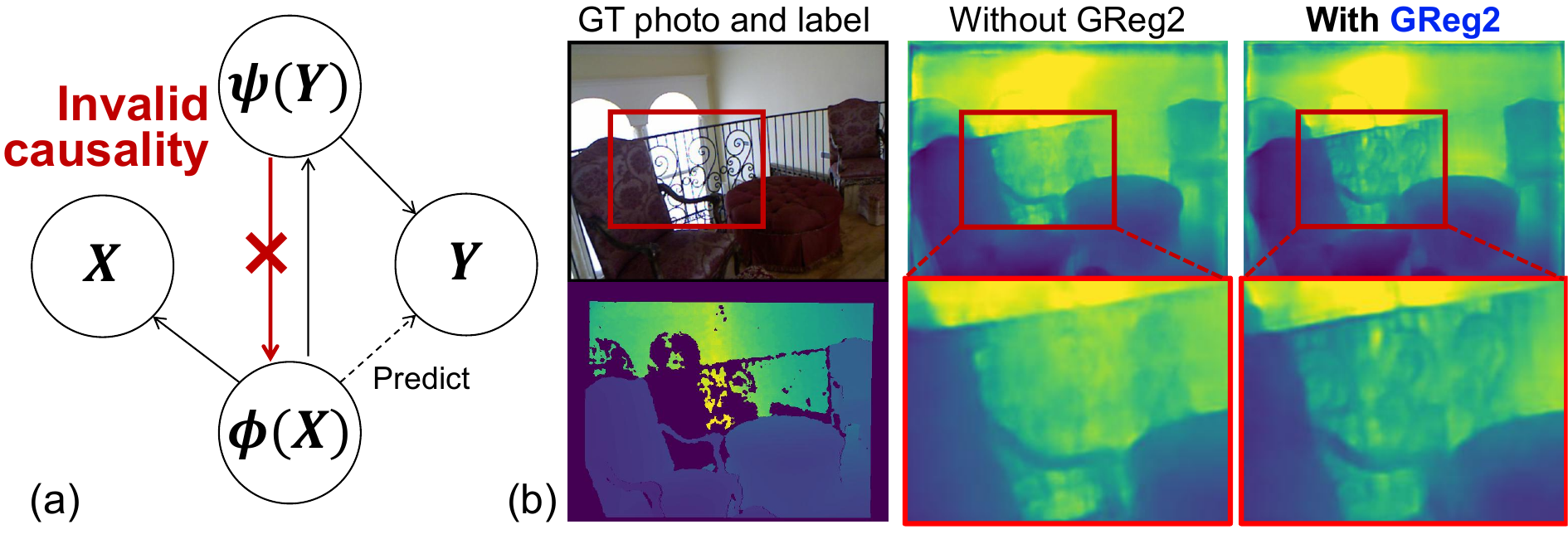}
\vspace{-5pt}
\caption{{(a)~Diagram of causality in the proposed method. (b)~Depth predictions on the unseen domain sample between model trained without and with \textbf{GReg2}.} }
\label{fig:cause}
\vspace{-8pt}
\end{figure}

Figure~\ref{fig:cause} graphically demonstrates the causal diagram under this scenario. 
{Since the prediction relationship from $\phi(\rmX) \to \rmY$ and the casual path $\psi(\rmY) \to \rmY$,  $\phi(\rmX) \to \psi(\rmY)$ should be preserved for prediction. However,  the casual path $\phi(\rmX) \to \psi(\rmY)$ may compromise the predication from $\phi(\rmX) \to \rmY$ when $\psi(\rmY)$ is unknown during the inference, leading to generalization degradation.}
{This unveils that the invalid causality from $\psi(\rmX)\to \phi(\rmY)$ that may happen during the learning invariance needs to be suppressed as $\max_{\phi,\psi} H(P(\phi(\rmX) | \psi(\rmY)), \mathcal{D})$ while $ \min _{\phi,\psi}H(P( \psi(\rmY) | \phi(\rmX) ), \mathcal{D})$, which can be simplified as: }
\begin{equation}
\label{eq:why_greg2}
        \min_{\phi,\psi} H (P(\phi(\rmX))) - H(P(\phi(\rmX))|P(\psi(\rmY))),
\end{equation}
where is \textbf{GReg2}.
See more mathematical details in Supplementary~\ref{app:our_obj}.
{Our experiments also unveil the phenomenon of invalid causality within invariant feature learning, where suppressing it could improve generalizability. 
The investigation of previous objectives also discloses that, in addressing the varying $P(\rmY|\mathcal{D})$, constructs akin to \textbf{GReg2} are often implicitly included (see Table~\ref{tab:summary}), though their efficacy was not explicitly stated.
Moreover, their efficacy may be compromised due to the lack of $\psi$ and other objective terms.}

Then, we assume that 
$\phi(X), \psi(Y)$ in the RKSH follow Multivariate Gaussian-like Distributions which are denoted as  $\mathcal{N}(\phi(\rmX); \mu_\rmX, \Sigma_{\rmX\rmX}), \mathcal{N}(\psi(\rmY); \mu_\rmY, \Sigma_{\rmY\rmY})$. 
$P(\phi(\rmX)  |  \psi(\rmY))$ follows $\mathcal{N}(\phi(\rmX) |  \psi(\rmY); \mu_{\rmX  |  \rmY}, \Sigma_{\rmX\rmX  |  \rmY})$.  \textbf{GReg2} can be simplified as: 
\begin{equation}
    \begin{split}
        \label{eq:cfs_detial}
    & H( \mathcal{N}(\phi(\rmX); \mu_\rmX, \Sigma_{\rmX\rmX}))  
    \\
    & - H( \mathcal{N}(\phi(\rmX)\mid \psi(\rmY); \mu_{\rmX|\rmY}, \Sigma_{\rmX\rmX|\rmY}))  \\
     = & \frac{1}{2} \ln(\frac{|\Sigma_{\rmX\rmX}|}{|\Sigma_{\rmX\rmX|\rmY}|}) \ge 0, 
    \end{split}
\end{equation}
where the inequality stands owing to the \textit{Condition Reducing Entropy}. This implies  $H( \mathcal{N}(\phi(\rmX); \mu_\rmX, \Sigma_{\rmX\rmX}))\!\! \ge\!\! H( \mathcal{N}(\phi(\rmX) \mid \psi(\rmY); \mu_{\rmX|\rmY}, \Sigma_{\rmX\rmX|\rmY}))$, deduced from $|\Sigma_{\rmX\rmX}| \!\!\ge\!\! |\Sigma_{\rmX\rmX|\rmY}|\!\! \ge \!\!0$, considering they are positive semi-definite. 
{Distinct from the \cite{yang2020causalvae,huang2022harnessing} which decompose causal effects through extra networks, our method is based on transfer entropy (TE) by ensuring $TE(\phi(\rmX) \!\!\!\to\!\!\! \psi(Y)) \!\!\!\ge\!\!\!TE(\phi(\rmX) \!\!\!\to\!\!\! \psi(Y))$, i.e., $H(\phi(\rmX)|\psi(Y))\!\!\! \ge \!\!\!H(\psi(Y)|\phi(\rmX)).$}
Thus, minimization of Eq.~\ref{eq:cfs_detial} occurs iff $|\Sigma_{\rmX\rmX}| =|\Sigma_{\rmX\rmX|\rmY}|$, reformulating the task as \!\!$min_{\phi,\psi} |\Sigma_{\rmX\rmX}|\!\!\! -\!\!\!|\Sigma_{\rmX\rmX|\rmY}|$, where $\Sigma_{\rmX\rmX \mid \rmY} = \Sigma_{\rmX\rmX} \!\!- \!\!\Sigma_{\rmX\rmY}\Sigma_{\rmY\rmY}^{-1}\Sigma_{\rmY\rmX}$, per~\citep{kay1993fundamentals}.
Therefore, 
\textbf{GReg2} is simplified as minimizing Conditional Feature Shift (CFS):
\begin{align}
\label{eq:x_shift}
    \min_{\phi,\psi}  |\Sigma_{\rmX\rmY}\Sigma_{\rmY\rmY}^{-1}\Sigma_{\rmY\rmX}|.
\end{align}


\subsection{Empirical losses derivations}
This section presents the empirical losses used to implement  Eq.~\ref{eq:dg_obj_final}. More detailed derivation can be referred to in Supplementary~\ref{app:our_obj}.
We introduce the mapping $\psi$ to relax the static target distribution. The implementation of $\psi$ varies across tasks, utilizing MLPs for classification and regression, and ResNet-50 for segmentation. To promote a consistent latent space, the mapped $\psi(\rmY)$ retains the same dimension as that of $\phi(\rmX)$. 
$\psi(\rmY)$ and $\phi(\rmX)$ are separately fed into $\mathcal{C}$ for making predictions and obtaining $\mathcal{L}_{A2}$ for posterior maximization:
\begin{align}
    \mathcal{L}_{A2}(\mathcal{C}, \phi,\psi) = H_c(\phi(\rmX), \rmY) +  H_c(\psi(\rmY), \rmY).
\end{align}
To mitigate domain shifts and learn domain invariance, we minimize cross-domain conditional feature distribution discrepancies. 
Specifically, the mean and variance of the joint distribution of $(\phi(\rmX), \psi(\rmY))$ in each domain are estimated using VAE encoders.
Consider $n$-pairs 
means and variance of $n$ domains, we derive a joint Gaussian distribution expression $P(\phi(\rmX_n),\psi(\rmY_n)) \triangleq \mathcal{N}(\vx_n, \vy_n;\mu_n, \Sigma_n)$. 
Accordingly, we establish $\mathbb{E}[ P ({\phi(\rmX_n), \psi(\rmY_n)})] \triangleq \mathcal{N}(\bar{\vx},\bar{\vy}; \bar{\mu}, \bar{\Sigma}
)$ where $\bar{\mu} = \mathbb{E}[\mu_n], \bar{\Sigma} = \mathbb{E}[\Sigma_n]$.
Base on PUB in Eq.~\ref{eq:PUB}, we introduce $\mathcal{L}_{A1}$ to minimize the conditional feature gap across domains:
\begin{align}
    \mathcal{L}_{A1}(\phi) 
    = \sum\nolimits_{i=1}^n
    (\log |\Sigma_i| + ||\bar{\mu} - \mu_i  ||^2_{\Sigma_i^{-1}}
    ).
\end{align}
To integrate prior information, similar to MIRO, we utilize VAE encoders to capture the means and variances of $\rmX$:
$P(\phi(\rmX)) \triangleq \mathcal{N}(\vx; \mu_x, \Sigma_x)$ and the output features $x_{\mathcal{O}}$ form $\mathcal{O}$. Given that $\mathcal{O}$ preserves the correlation between $\rmX$ ($\phi(\rmX)$) and $\rmY$ ($\psi(\rmY)$), and is frozen during training, $\rmY, \psi(\rmY)$ is omitted in empirical loss.
We propose $\mathcal{L}_{R1}$ to minimize the divergence between features and $\mathcal{O}$:
\begin{align}
    \mathcal{L}_{R1}(\phi) 
    = \log |\Sigma_x| + ||x_{\mathcal{O}} - \mu_x  ||^2_{\Sigma_x^{-1}}.
\end{align}


For suppressing the invalid causality, derived from Eq.~\ref{eq:x_shift}, the loss is designed to minimize the CFS:
\begin{align}
\label{eq:L_CFS}
        \mathcal{L}_{R2}(\phi) = ||\Sigma_{\rmX\rmY}\Sigma_{\rmY\rmY}^{-1}\Sigma_{\rmY\rmX}||_2 ,
\end{align} 
where 
$\Sigma_{\rmX\rmY}= 
\mathbb{E}[(\phi(\rmX)-\mathbb{E}[\phi(\rmX)])^{\top} (\phi(\rmY)-\mathbb{E}[\phi(\rmY)])]$, and a similar calculation process is done for $\Sigma_{\rmY\rmY}$ and $\Sigma_{\rmY\rmX}$. The final loss is a weighted combination of the above losses\footnote{See detailed hyper-parameters settings in Supplementary~\ref{app:para}.}:
\begin{align}
    \mathcal{L}(\mathcal{C}, \phi, \psi) \!\!  =   \!\! v_{A1}\mathcal{L}_{A1} \!\! + \!\!  v_{A2} \mathcal{L}_{A2} \!\!+\!\! v_{R1}\mathcal{L}_{R1} \!\! + \!\!  v_{R2}\mathcal{L}_{R2}.
\end{align}




\section{{Connection to previous methods}}
\label{sec:conn}
We validate our objective function's efficiency theoretically and demonstrate its connections with previous objectives, indicating that previous mDG efforts have partly optimized the proposed objective. Refer to Table~\ref{tab:summary} and Supplementary~\ref{app:pre_obj} for a detailed understanding of previous objectives.

\textbf{Using $\psi$ \textit{v.s.} not using $\psi$.}
Previous works rarely employed $\psi$ to map $\rmY$, whereas we show its benefits for mDG tasks. 
Employing \textit{Jensen's inequality}, we obtain $H(\mathbb{E}[ P ({\phi(\rmX_n), \psi(\rmY_n)})]) \ge H(\mathbb{E}[ P ({\phi(\rmX_n), \rmY_n})])$. 
When other objectives remain the same, 
we compare the model with parameters $\theta^{\psi}$ optimized via the $\psi$ mapping, against another model without $\psi$ using parameters $\theta^{n\psi}$:
\begin{align}
\label{eq:risk_compare_psi}
    \sup R({\theta}^{n\psi}) \ge \sup R({\theta}^{\psi}).
\end{align}
The equivalence is valid only if $\psi$ serves as a bijection, a condition prevalent in practical scenarios like classification. 
Thus, this mapping does not hinder model performance in classification tasks. 
It also implies that using $\psi(\rmY)$ can lower generalization risks after optimization, especially when $\rmY$ contains features dependent on $\mathcal{D}$. This could potentially yield superior generalization in segmentation and regression tasks. Detailed proof can be seen in Supplementary~\ref{app:our_obj}. 

\begin{remark}[Importance of $\rmY$ mapping $\psi$] 
    Besides relaxing the static distribution assumption of $\rmY$, $\psi$ conveys two other notable benefits:
    1). $\rmX$ and $\rmY$ may originate from different sample spaces with distinct shapes. By applying mappings, $\psi(\rmY)$  can be adapted to the same shape as $\phi(\rmX)$. In practice, concatenating $\phi(\rmX)$ and $\psi(\rmY)$ is often used as input for VAE encoders to capture $P ({\psi(\rmY)}, \phi(\rmX))$.
    2). The derivation of Eq.~\ref{eq:x_shift} requires the computation of covariance, which mandates that two variables occupy the same sample space - a condition fulfilled by applying $\psi(\rmY)$. 
\end{remark}

\textbf{{Incorporating conditions leads to lower generalization risk on learning invariant representations.}}
A few past works~\citep{ganin2016domain,sun2016deep} minimize domain gaps between features without condition consideration. Its objective for \textbf{Aim1} is:
\begin{equation}
 \label{eq:cond_t1}
    \resizebox{.88\hsize}{!}{$
    \begin{split}
    H(P ({\phi(\rmX)\mid \mathcal{D}}))  
    \le & H( P ({\phi(\rmX)\mid \mathcal{D}})) + \\ & H( P (\psi(\rmY)  \mid \phi(\rmX)),\mathcal{D})
    = \text{\textbf{GAim1}}.  
    \end{split}
    $}
\end{equation}
While the other objectives are identical, we consider a model with parameters ${\theta}^{nc}$, trained with $\min_{\psi} H( P ({\phi(\rmX_n)}))$, against another model with ${\theta}^{c}$ parameters, trained with \textbf{GAim1}. In this scenario, their empirical risks satisfy:  
\begin{align}
\label{eq:risk_compare_cond}
    \sup R({\theta}^{nc}) \ge \sup R({\theta}^{c}).
\end{align}
See the mathematical details in Supplementary~\ref{app:our_obj}.
This reveals that without condition consideration, the minimization of generalization risk is merely partial due to the overlooked risk correlated to $\rmY$. Additional evidence supporting the importance of condition consideration is provided by CDANN~\citep{li2018deep} and CIDG~\citep{li2018domain}.
Our experiments, conducted through a uniform implementation, also lend support to it.
 

\textbf{Effect of oracle model $\mathcal{O}$.} 
As stated by MIRO~\citep{cha2022miro} and SIMPLE~\citep{li2022simple},  a generalized $\mathcal{O}$ comprising both seen and unseen domains yields significant improvements. During the derivation of Eq.~\ref{eq:PUB}, we find that the disregard \textbf{GAim1} term in  MIRO~\citep{cha2022miro} and SIMPLE~\citep{li2022simple} may result in inferior outcomes to our proposed objective.

\begin{remark}[Synergy of learning invariance, integrating prior knowledge and suppressing invalid causally] For readability, we have divided the overall mDG objective into four aspects despite all terms being interconnected.
Specifically, as shown by PUB in Eq.~\ref{eq:PUB}, \textbf{GReg1} collaborating with \textbf{GAim1} brings more performance gains than the case when it is solely applied. 
{Moreover, Eq.~\ref{eq:GAim1_GReg2} shows that the side effect of invalid causality in \textbf{GAim1} is alleviated by combining with \textbf{GReg2}}, underscoring the significance of combining learning invariance, integrating prior knowledge, and suppressing invalid causally.
It also suggests that all terms are synergistic and contribute together to improved results.
\end{remark}

Validating our assertions via experiments, \Secref{sec:res_ablation} ablation study finds that simple cross-domain covariance limitation (\textbf{GReg2}) cannot ensure improved results with prior knowledge.

\section{Experiments}
\label{sec:exp}
Four groups of experiments are done to validate the proposed GMDG.
A toy example validates the relaxation of $P(\rmY|\mathcal{D})$ static assumption brought by $\psi$ of $\rmY$. 
Furthermore, 
we conduct experiments on regression, segmentation, and classification tasks and use complex benchmark datasets.
For a simplification, please refer to Table~\ref{tab:note} for the formulations of terms and their alias.

\subsection{Toy experiments on synthetic datasets}
We perform a regression task on synthetic data to illustrate the impact of using $\psi$, showcasing its potential for superior results if $\psi$ is not bijective. 

\begin{table}[t]
\centering
\resizebox{\columnwidth}{!}{%
     \begin{tabular}{ccccccc}
     \\
    \toprule
         &  \multicolumn{3}{c}{Affine transformations} &  \multicolumn{3}{c}{Squared and cubed transformations}\\ \midrule
            \multicolumn{1}{c}{}& ERM & +$\mathcal{L}_{A1}(\phi)$ & \multicolumn{1}{c}{\cellcolor{mygray} +$\mathcal{L}_{A1}(\phi,\psi)$} & ERM & +$\mathcal{L}_{A1}(\phi)$ & \multicolumn{1}{c}{\cellcolor{mygray}+$\mathcal{L}_{A1}(\phi,\psi)$}  \\
           {No DCDS} & 0.3485 & 0.3537 & \cellcolor{mygray}\textbf{0.3369}  & 1.5150 & 0.4652 & \cellcolor{mygray}\textbf{0.3370} \\
           {With DCDS} & 0.4144 & 0.2290 & \cellcolor{mygray}\textbf{0.1777} & 0.8720 & 1.5868 & \cellcolor{mygray}\textbf{0.8241} \\ \bottomrule
    \end{tabular} %
    }
\vspace{-5pt}
\caption{Toy experimental results: MSE losses on testing set.+$\mathcal{L}_{A1}(\phi)$ denotes $\mathcal{L}_{A1}$ is used without $\psi$ while +$\mathcal{L}_{A1}(\phi,\psi)$ denotes $\psi$ is used. Best results are highlighted as \textbf{bold}. DCDS denotes domain-conditioned distribution shift.}
 \label{tab:toy_res}
 \vspace{-5pt}
\end{table}
\begin{table}[t]
    \centering
\resizebox{\columnwidth}{!}{%
\begin{tabular}{ll|ll}
\toprule
\textbf{GAim2} & $H(P ({\psi(\rmY) \mid  \phi(\rmX)})) + H(P ({\rmY \mid  \psi(\rmY)}))$ & 
\textbf{GReg1} & $\KL(P ({\phi(\rmX), \rmY} \mid \mathcal{D}) \Vert \mathcal{O}))$  \\ 
\textbf{iAim1} & $H( P ({\phi(\rmX)} \mid \mathcal{D}))$ & 
\textbf{GAim1} & $ H(  P ({\phi(\rmX), \rmY}) \mid \mathcal{D})$  \\
\textbf{iReg2} & $ -H( P ({\phi(\rmX)} , \mathcal{D}) + H(P({\phi(\rmX)}))$ & 
\textbf{GReg2} & ${ - H(P(\phi(\rmX)  \mid  \psi(\rmY))) 
+ H(P(\phi(\rmX)))}$ \\  
\bottomrule
\end{tabular}%
}
\vspace{-5pt}
\caption{Notations for terms.}
\label{tab:note}
\vspace{-13pt}
\end{table}

\begin{figure*}[t]
    \centering
    \includegraphics[width=\linewidth]{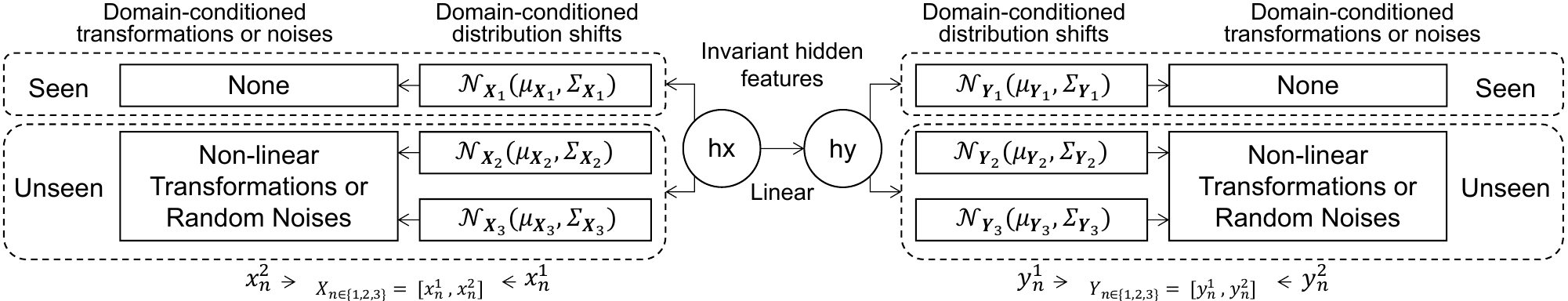}
    \caption{Toy experiments: Diagram of constructing the toy dataset. 
    }
    \label{fig:toy_figure}
\end{figure*}

\textbf{Synthetic data.} 
{Supplementary~\Figref{fig:toy_figure}} illustrates the construction of synthetic data, built on $\rmX$-$\rmY$ pair latent features with a linear relationship, ensuring invariant existence.  
To better explore this issue, we created four distinct data groups: without and with distribution shift, used affine or squared and cubed transformations as domain-conditioned transformations, and their cross combinations. More description can be seen in Supplementary~\ref{app:para}.

\textbf{Experimental setup.} We use two of three constructed domains for training and validation and the last one for testing. Validation and test losses are calculated by MSE. 
To maintain fairness, all experiments adopt the same network which is selected by the best validation results.  
Learning aims to find invariant hidden features of $X, Y$ while preserving predictive ability from unseen $X$ to $Y$.

\textbf{Results.} Toy experiment results are reported in Table~\ref{tab:toy_res}, which are also visualized in \Figref{fig:toy_res}. It is observed that across all settings,  employing $\psi$ with $\mathcal{L}_{A1}$ yields superior results, outperforming ERM and ERM+$\mathcal{L}_{A1}$($\phi$) without $\psi$, validating the enhanced generalization effect brought by utilizing $\psi$ whenever $Y$ varies per domain, supporting Eq.~\ref{eq:risk_compare_psi}. 
Supplementary~\Figref{fig:toy_res}   shows that $\psi$ does learn the invariant representations for $Y$ to relax previous $Y$-invariant assumption.  
Specifically, learning the invariance of $Y$ with $\psi$ results in superior invariant representations as the latent representations of $X,Y$ are primarily linear,  aligning with $X$ and $Y$'s linear relationship during data construction.
The bottom-left figures reveal that though ERM has learned the most invariant $\phi(X)$, it suffers the worst test loss, indicating that a well-learned invariant $\phi(X)$ is not sufficient when $Y$ also has domain-dependent traits. The results also suggest that assuming that $Y$ vary 
across domains, using $\mathcal{L}_{A1}$ without $\psi$ may not yield superior results. 

\begin{table}[t]
\centering
\resizebox{\columnwidth}{!}{%
\begin{tabular}{l|ccccc|cc|p{0.3cm}}
\toprule
                              & SILog$\downarrow$   & Abs Rel$\downarrow$ & RMS$\downarrow$    & Sq Rel$\downarrow$ & RMS log$\downarrow$ & $\delta_{1}$$\uparrow$    & $\delta_{2}$$\uparrow$     & TD           \\ \midrule
Backbone (Swin-L)
& 11.1473          & 10.98          & 56.11        & 8.86          & 14.32          & 87.47          & 98.05          &     \\
VA-DepthNet (\textbf{GAim2})
& 10.9357          & 11.15          & 56.36          & 9.02           & 14.41          & 87.73          & 98.02          & \multirow{4}{*}{S}    \\
\blue{\textbf{GReg1}}+\textbf{GAim2}
& 10.6548          & 10.49          & 52.63          & 8.04           & 13.72          & 89.75          & 98.12\\
\blue{\textbf{GAim1}}+\blue{\textbf{GReg1}}+\textbf{GAim2}
& 10.1924          & 10.39          & \textbf{50.52} & \textbf{7.68}  & 13.39          & 89.86          & \textbf{98.17}\\
\rowcolor{mygray}\begin{tabular}[c]{@{}l@{}}\blue{\textbf{GAim1}}+\blue{\textbf{GReg1}}+\\ \textbf{GAim2}+\blue{\textbf{GReg2}} (GMDG)\end{tabular} 
& \textbf{10.1402} & \textbf{10.27} & 50.59          & 7.72           & \textbf{13.22} & \textbf{90.53} & 97.98 \\ \midrule
Backbone (Swin-L)
& 14.2078          & 16.20          & 81.22         & 16.30          & 20.59          &  72.44         & 96.35          &     \\
VA-DepthNet (\textbf{GAim2})
& 14.7080          & 16.76          & 83.17          & 17.07          & 21.44          & 71.46          & 95.11          & \multirow{4}{*}{Co}\\
\blue{\textbf{GReg1}}+\textbf{GAim2}
& 14.1600          & 16.41          & 80.78          & 16.56          & 21.02          & 72.06          & 95.38 \\
\blue{\textbf{GAim1}}+\blue{\textbf{GReg1}}+\textbf{GAim2}
& \textbf{13.9978} & 15.90          & 77.97          & 15.70          & 20.25          & 73.37          & \textbf{95.86} \\
\rowcolor{mygray}\begin{tabular}[c]{@{}l@{}}\blue{\textbf{GAim1}}+\blue{\textbf{GReg1}}+\\ \textbf{GAim2}+\blue{\textbf{GReg2}} (GMDG)\end{tabular} 
& 14.2803          & \textbf{15.57} & \textbf{77.45} & \textbf{15.27} & \textbf{19.94} & \textbf{74.40} & 95.47 \\  \midrule
Backbone (Swin-L)
& 11.6132          & 12.87          & 44.51         & 8.01          & 15.58          & 84.57          & 97.87          &     \\
VA-DepthNet (\textbf{GAim2})
& 11.5080          & 12.50          & 43.98          & 7.67           & 15.37          & 84.78          & 97.87           & \multirow{4}{*}{O}\\
\blue{\textbf{GReg1}}+\textbf{GAim2}
& 10.4061          & 11.71          & 39.02          & 6.87           & 13.83          & 88.27          & 98.17\\
\blue{\textbf{GAim1}}+\blue{\textbf{GReg1}}+\textbf{GAim2}
& 10.4907          & 11.53          & \textbf{38.43} & 6.69           & 13.68          & 88.46          & \textbf{98.17} \\
\rowcolor{mygray}\begin{tabular}[c]{@{}l@{}}\blue{\textbf{GAim1}}+\blue{\textbf{GReg1}}+\\ \textbf{GAim2}+\blue{\textbf{GReg2}} (GMDG)\end{tabular} 
& \textbf{10.4438} & \textbf{11.33} & 38.95          & \textbf{6.66}  & \textbf{13.67} & \textbf{88.86} & 98.16 \\ \midrule
Backbone (Swin-L)
& 14.7350          & 18.36          & 52.31          & 13.05          & 20.06          & 74.74          & 93.94          &     \\
VA-DepthNet (\textbf{GAim2})
& 15.0300          & 17.99          & 56.54          & 13.20          & 20.64          & 72.40          & 94.38          & \multirow{4}{*}{\textbf{H}}\\
\blue{\textbf{GReg1}}+\textbf{GAim2}
& 14.7377          & 17.02          & 55.39          & 12.06          & 19.86          & 74.05          & 95.17\\
\blue{\textbf{GAim1}}+\blue{\textbf{GReg1}}+\textbf{GAim2}
& 14.5018          & 17.14          & \textbf{52.10} & 12.01          & 19.37          & 76.13          & 94.95\\
\rowcolor{mygray}\begin{tabular}[c]{@{}l@{}}\blue{\textbf{GAim1}}+\blue{\textbf{GReg1}}+\\ \textbf{GAim2}+\blue{\textbf{GReg2}} (GMDG)\end{tabular}
& \textbf{14.1414} & \textbf{15.90} & 52.22          & \textbf{10.72} & \textbf{18.95} & \textbf{76.27} & \textbf{96.10} \\ \midrule
Backbone (Swin-L)
& 12.9258         &  14.60         &  58.54        & 11.56         & 17.64          & 79.81          & 96.55          & \\
VA-DepthNet(\textbf{GAim2}) 
& 13.0454          & 14.60          & 60.01          & 11.74          & 17.97          & 79.09          & 96.35          & \multirow{4}{*}{Avg.}\\ 
\blue{\textbf{GReg1}}+\textbf{GAim2}
& 12.4897          & 13.91          & 56.96          & 10.88          & 17.11          & 81.03          & 96.71 \\
\blue{\textbf{GAim1}}+\blue{\textbf{GReg1}}+\textbf{GAim2}
& 12.2957          & 13.74          & \textbf{54.76}          & 10.52          & 16.67          & 81.96          & 96.79  \\
\rowcolor{mygray}\begin{tabular}[c]{@{}l@{}}\blue{\textbf{GAim1}}+\blue{\textbf{GReg1}}+\\ \textbf{GAim2}+\blue{\textbf{GReg2}} (GMDG)\end{tabular} 
& \textbf{12.2514} & \textbf{13.27} & {54.80} & \textbf{10.09} & \textbf{16.45} & \textbf{82.52} & \textbf{96.93}                       
\\
\bottomrule
\end{tabular}%
}
\vspace{-5pt}
\caption{Regression results: Comparison of results between proposed and previous methods. Added terms to the baseline are highlighted as \blue{\textbf{blue}}. The best results for each group are highlighted in \textbf{bold}.  TD: Test Domain. 
} 
\label{tab:reg_res}
\vspace{-13pt}
\end{table}

\subsection{Regression on benchmark datasets: Monocular depth estimation}
We conduct the Monocular Depth Estimation task as the real-world regression task to further verify GMDG.

\textbf{Experimental setup.}  We employ VA-DepthNet~\cite{liu2023va} with Swin-L~\cite{Liu_2021_ICCV} backbone as the baseline and follow their hyperparameter settings. Experiments are conducted on NYU Depth V2~\cite{silberman2012indoor}. To construct multiple domains, we split the dataset into four categories: `School' (S), `Office' (O), `Home' (H), and `Commercial' (Co).
We conduct the standard leave-one-out cross-validation as an evaluation method. 
We use the best checkpoint on the seen domains for the evaluation. 
{Note that all models are trained on the newly constructed dataset.}
Statistical results on popular evaluation metrics such as the square root of the Scale Invariant Logarithmic error (SILog), Relative Squared error (Sq Rel), Relative Absolute
Error (Abs Rel), Root Mean Squared error (RMS), and threshold accuracy ($\delta_{1},\delta_{2}$) are used as evaluation metrics.
See more experimental details in the Supplementary~\ref{app:para}.  


\textbf{Results.} 
{{The Monocular Depth Estimation results are exhibited in Table~\ref{tab:reg_res}. 
It can be seen that using terms that are proposed in GMDG leads to better generalization on unseen domains, and 
using the full GMDG leads to the best results in most metrics. The improvements suggest the feasibility of our GMDG in real-world regression tasks.
Specifically, using \textbf{GReg2} with other terms significantly improves the results when Home is the unseen domain, {which is the most difficult domain to be generalized for the VA-DepthNet,} and barely compromises the performances while using other domains as the unseen. This reveals that suppressing the causality can improve the generalization of the model (refer to Figure~\ref{fig:cause}~(b) for visual results). Note that, except SILog, all metrics results are scaled by $100$ for readability; due to the lack of objective targeting on the mDG problem, VA-DepthNet performs worse than its baseline.  
See more visualizations in Supplementary~\ref{app:res}.}}

\subsection{Segmentation on benchmark datasets}
\label{sec:seg_task}

\textbf{Experimental setup.}
We follow the experimental setup of RobustNet~\citep{choi2021robustnet} for mDG segmentation experiments, particularly using DeepLabV3+~\citep{chen2018encoder} as the semantic segmentation model architecture, with ResNet-50 backbone and SGD optimizer. 
As shown in Table~\ref{tab:summary}, RobustNet's objective is equivalent to using \textbf{GAim2} and \textbf{GReg2}. 
Consistent with previous methods, mIoU serves as our evaluation metric.
Datasets comprise real-world datasets (Cityscapes~\citep{cordts2016cityscapes} (Ci), BDD-100K~\citep{yu2020bdd100k} (B),
Mapillary~\citep{neuhold2017mapillary} (M)) and synthetic datasets (GTAV~\citep{richter2016gtav}, SYNTHIA~\citep{ros2016synthia}). Specifically, we train a model on GTAV and Cityscapes, testing on other datasets.  We compare our results to DeepLabv3+\cite{chen2018encoder}, IBN-NET\cite{pan2018two} and RobustNet~\citep{choi2021robustnet}. We use Intersection over Union (mIoU) as the evaluation metric. See Supplementary~\ref{app:para} for more experimental details.

\begin{table}[t]
    \centering
    \resizebox{0.99\columnwidth}{!}{%
    \centering
    \begin{tabular}{lccc|c}
    \\
    \toprule
    TD & \textbf{Ci} & \textbf{B} & \textbf{M} & \textbf{Avg.} \\ \midrule
    DeepLabv3+ & 35.46 & 25.09 & 31.94 & 30.83 \\
    IBN-Net & 35.55 & 32.18 & 38.09  & 35.27 \\ \midrule
    RobustNet (\textbf{GAim2}, \textbf{GReg2}) & 37.69 & 34.09 & 38.49 & 36.76 \\
     \blue{\textbf{GAim1}}+\textbf{GAim2}+\textbf{GReg2} & 38.58 &  {34.72} &  39.11 &  37.47 \\
    \blue{\textbf{GReg1}}+\textbf{GAim2}+\textbf{GReg2} & 38.13  &  \textbf{35.02} &  39.29 & 37.48\\
     \rowcolor{mygray}\blue{\textbf{GAim1}}+\blue{\textbf{GReg1}}+\textbf{GAim2}+\textbf{GReg2} (GMDG)& \textbf{38.62} & 34.71 & \textbf{39.63} & \textbf{37.65}
     \\
     \bottomrule
    \end{tabular}%
    }
    \vspace{-5pt}
    \caption{Segmentation results: Comparison of mIoU($\%$) between proposed and previous methods. The models are trained on GTAV and SYNTHIA domains. 
    The added objective terms are highlighted as \blue{\textbf{blue}}. The best results are highlighted in \textbf{bold}. }
    \label{tab:seg_res}
\vspace{-15pt}
\end{table}

\textbf{Results.} 
Table~\ref{tab:seg_res} shows the efficacy of our proposed objective in segmentation tasks upon introducing $\psi$. 
Ablation results highlight that using $\psi$ alongside \textbf{GAim1} can enhance baseline performance, experimentally substantiating that the introduction of $\psi$, in relaxing assumptions, boosts performance for better generalization. Using \textbf{GReg1} alone also improves average mIoU. Importantly, the most enhancement in average mIoU is observed when \textbf{GReg1} and \textbf{GAim1} are used together, which finds validation in the PUB derivation in Eq.~\ref{eq:PUB}. See more results and visualizations in Supplementary~\ref{app:res}.

\subsection{Classification on benchmark datasets}
\label{sec:exp_real}
\textbf{Experimental setup.} 
We operate on the DomainBed suite~\cite {gulrajani2020search} and leverage standard leave-one-out cross-validation as an evaluation method.
We experiment on $5$ real-world benchmark datasets, including PACS~\citep{li2017deeper}, VLCS~\citep{fang2013unbiased}, OfficeHome~\citep{venkateswara2017deep}, TerraIncognita~\citep{beery2018recognition}, and DomainNet~\citep{peng2019moment}.
The results are the averages from three trials of each experiment. Following MIRO, two backbones are used for the training 
(ResNet-50~\citep{he2016deep} 
pre-trained in the ImageNet~\citep{he2016deep} and RegNetY-16GF backbone with SWAG pre-training~\citep{singh2022revisiting}). 
The backbones are trained with our proposed objective barely and further with SWAD~\citep{cha2021swad}, respectively. 
See Supplementary~\ref{app:para} for more experimental details.

\begin{table}[t]
\resizebox{\columnwidth}{!}{%
\begin{tabular}{lccccc|c}
 \toprule
& \multicolumn{6}{c}{\bf Non-ensemble methods} \\ 
 TD & \bf PACS & \bf VLCS & \bf OfficeHome & \bf TerraInc & \bf DomainNet & \bf Avg. \\ \midrule
MMD~\citep{li2018domainMMD} & 84.7±0.5 & 77.5±0.9 & 66.3±0.1 & 42.2±1.6 & 23.4±9.5 & 58.8 \\
Mixstyle~\citep{zhou2021domain} & 85.2±0.3 & 77.9±0.5 & 60.4±0.3 & 44.0±0.7 & 34.0±0.1 & 60.3 \\
GroupDRO~\citep{sagawa2019distributionally} & 84.4±0.8 & 76.7±0.6 & 66.0±0.7 & 43.2±1.1 & 33.3±0.2 & 60.7 \\
IRM~\cite{arjovsky2019invariant} & 83.5±0.8 & 78.5±0.5 & 64.3±2.2 & 47.6±0.8 & 33.9±2.8 & 61.6 \\
ARM~\citep{zhang2021adaptive} & 85.1±0.4 & 77.6±0.3 & 64.8±0.3 & 45.5±0.3 & 35.5±0.2 & 61.7 \\
VREx~\citep{krueger2021out} & 84.9±0.6 & 78.3±0.2 & 66.4±0.6 & 46.4±0.6 & 33.6±2.9 & 61.9 \\
CDANN~\citep{li2018deep} & 82.6±0.9 & 77.5±0.1 & 65.8±1.3 & 45.8±1.6 & 38.3±0.3 & 62.0 \\
DANN~\citep{ganin2016domain} & 83.6±0.4 & 78.6±0.4 & 65.9±0.6 & 46.7±0.5 & 38.3±0.1 & 62.6 \\
RSC~\citep{huang2020self} & 85.2±0.9 & 77.1±0.5 & 65.5±0.9 & 46.6±1.0 & 38.9±0.5 & 62.7 \\
MTL~\citep{blanchard2021domain} & 84.6±0.5 & 77.2±0.4 & 66.4±0.5 & 45.6±1.2 & 40.6±0.1 & 62.9 \\
MLDG~\citep{li2018learning} & 84.9±1.0 & 77.2±0.4 & 66.8±0.6 & 47.7±0.9 & 41.2±0.1 & 63.6 \\
Fish~\citep{shi2021gradient} & 85.5±0.3 & 77.8±0.3 & 68.6±0.4 & 45.1±1.3 & 42.7±0.2 & 63.9 \\
ERM~\citep{Vapnik1998ERM} & 84.2±0.1 & 77.3±0.1 & 67.6±0.2 & 47.8±0.6 & 44.0±0.1 & 64.2 \\
SagNet~\citep{nam2021reducing} & {86.3}±0.2 & 77.8±0.5 & 68.1±0.1 & 48.6±1.0 & 40.3±0.1 & 64.2 \\
SelfReg~\citep{kim2021selfreg} & 85.6±0.4 & 77.8±0.9 & 67.9±0.7 & 47.0±0.3 & 42.8±0.0 & 64.2 \\
CORAL~\citep{sun2016deep} & 86.2±0.3 & 78.8±0.6 & 68.7±0.3 & 47.6±1.0 & 41.5±0.1 & 64.5 \\
mDSDI~\cite{bui2021exploiting} & 86.2±0.2 & 79.0±0.3 & 69.2±0.4 & 48.1±1.4 & 42.8±0.1 & 65.1 \\ 

 & \multicolumn{6}{c}{Use ResNet-50~\citep{he2016deep} as oracle model.} \\ 
Style Neophile~\cite{kang2022style} & \textbf{89.11} & - & 65.89 & - & 44.60 & - \\
MIRO~\citep{cha2022miro} (\textbf{GReg1}) & 85.4±0.4 & 79.0±0.3 & 70.5±0.4 & 50.4±1.1 & 44.3±0.2 & 65.9 \\ 
\rowcolor{mygray}\blue{\textbf{GMDG}} & 85.6±0.3 & \textbf{79.2}±0.3 &\textbf{70.7}±0.2 & \textbf{51.1}±0.9 & \textbf{44.6}±0.1 & \textbf{66.3} \\ \midrule
 & \multicolumn{6}{c}{Use RegNetY-16GF~\citep{singh2022revisiting} as oracle model.} \\ 
MIRO  & \textbf{97.4}±0.2 & 79.9±0.6 & 80.4±0.2 & 58.9±1.3 & 53.8±0.1 & 74.1 \\
\rowcolor{mygray}\blue{\textbf{GMDG}}  & 97.3±0.1 & \textbf{82.4}±0.6 & \textbf{80.8}±0.6 & \textbf{60.7}±1.8 & \textbf{54.6}±0.1 & \textbf{75.1} \\ \bottomrule \toprule
& \multicolumn{6}{c}{\bf Ensemble methods} \\ 
 & \bf PACS & \bf VLCS & \bf OfficeHome & \bf TerraInc & \bf DomainNet & \bf Avg. \\ \hline
& \multicolumn{6}{c}{ Use multiple oracle models.} \\ 
SIMPLE~\citep{li2022simple} & 88.6±0.4 & 79.9±0.5 & 84.6±0.5 & 57.6±0.8 & 49.2±1.1 & 72.0 \\
SIMPLE++~\citep{li2022simple} & \textbf{99.0}±0.1 & \textbf{82.7}±0.4 & \textbf{87.7}±0.4 & \textbf{59.0}±0.6 & \textbf{61.9}±0.5 & \textbf{{78.1}} \\ \midrule 
& \multicolumn{6}{c}{Use ResNet-50~\citep{he2016deep} as oracle model.} \\ 
MIRO + SWAD  &  \textbf{88.4}±0.1 & {{79.6}}±0.2 & {{72.4}}±0.1 & {{52.9}}±0.2 & {{47.0}}±0.0 & {{68.1}} \\
\rowcolor{mygray}\begin{tabular}[c]{@{}l@{}} \blue{\textbf{GMDG}} + SWAD \end{tabular} & \textbf{{88.4}}±0.1 & \textbf{79.6}±0.1 & \textbf{72.5}±0.2 & \textbf{53.0}±0.7 & \textbf{47.3}±0.1 &  \textbf{68.2} \\ 
\hline
& \multicolumn{6}{c}{Use RegNetY-16GF~\citep{singh2022revisiting} as oracle model.} \\ 
MIRO + SWAD  & 96.8±0.2 & 81.7±0.1 & 83.3±0.1 & {{64.3}}±0.3 & 60.7±0.0 & 77.3 \\
\rowcolor{mygray}\blue{\textbf{GMDG}} + SWAD
& \textbf{{97.9}}±0.3 & \textbf{{82.2}}±0.3 & \textbf{{84.7}}±0.2 & \textbf{65.0}±0.2 & \textbf{{61.3}}±0.2 & \textbf{78.2} \\ 
\bottomrule
\end{tabular}%
}
\vspace{-5pt}
\caption{Classification results: Comparison of results between the proposed and previous non-ensemble and ensemble mDG methods. The best results for each group are highlighted in \textbf{bold}.}
\label{tab:main_res1}
\vspace{-13pt}
\end{table}

\textbf{Results.} 
Table~\ref{tab:main_res1} displays the results of non-ensemble algorithms and ensemble algorithms that employ pre-trained models as oracle models.
Specifically, our proposed objectives demonstrate more substantial improvements when a higher-quality pre-trained oracle model ($\mathcal{O}$) is applied. When employing the ResNet-50 model, our approach yields average improvements of approximately $0.3\%$ and $0.1\%$ without and with SWAD, respectively, compared to MIRO. In contrast, when RegNetY-16GF serves as an oracle, GMDG results in significant average improvements of $1.1\%$ and $0.9\%$  without and with SWAD, respectively. 
Remarkably, our approach outperforms $0.1\%$ more than the SOTA method, SIMPLE++, which relies on an ensemble of 283 pre-trained models as oracle models, whereas ours only engages a single pre-trained model.
Overall, these results strongly support GMDG's effectiveness in classification tasks. See more results in Supplementary~\ref{app:res}.

\subsection{Ablation studies}
\label{sec:res_ablation}

To better compare our objective with previous objectives, we conduct a systematic ablation study on the classification task since most previous objectives are only available for classification due to the lack of $\psi$. 
\textbf{Experimental setup.} 
In the ablation studies, we test varied terms (see Table~\ref{tab:abl_res}) combinations on the HomeOffice dataset using SWAG pre-training~\citep{singh2022revisiting} and SWAD~\citep{cha2021swad}.
Every experiment is repeated in three trials, 
sharing the same hyper-parameter settings for evaluation. 
See Supplementary~\ref{app:para} for more details.

\textbf{Results.} Table~\ref{tab:abl_res} presents ablation study results. The first column denotes previous methods equivalent to term combinations.  
The main findings are as follows. See Supplementary~\ref{app:other_findings} for more other findings.

\begin{table}[t]
\centering
\resizebox{\columnwidth}{!}{%
\begin{tabular}{p{5.2
cm}ccccc|c}
\toprule
 \textbf{Used objectives}& \textbf{Art} & \textbf{Clipart} & \textbf{Product} & \textbf{Real} & \textbf{Avg.} & \textbf{Imp.}  \\   \midrule
 & \multicolumn{5}{c}{Without $\mathcal{O}$ (\textbf{GReg1})} \\
\textbf{GAim2} (ERM) & 78.4±0.7 & 68.3±0.5 & 85.8±0.4 & 85.8±0.3 & 79.6±0.2 & 0.0\\ 
\textbf{GAim2 + iAim1} (DANN) & 79.1±1.0 & 68.6±0.0 & 85.6±0.8 & 86.1±0.5 & 79.8±0.2 & +0.2\\
\textbf{GAim2 + GAim1} (CDANN, CIDG) & 79.1±0.7 & 69.1±0.1 & 85.7±0.5 & 86.3±0.6 & 79.9±0.4 & +0.3\\
\textbf{GAim2  +iReg2} (CORAL+$\psi$) & 79.1±0.1 & 69.9±0.4 & 86.0±0.1 & 86.3±0.4 & 80.3±0.2 & {+0.7}\\
\textbf{GAim2 + GReg2}  & 79.2±0.1 & \textbf{69.9}±1.4 & 86.1±0.5 & 86.1±0.1 & 80.3±0.3 & +0.7\\
\rowcolor{mygray} 
\begin{tabular}[c]{@{}l@{}}\textbf{GAim2 + GAim1} \\ \textbf{+ GReg2} (MDA+$\psi$)\end{tabular}
& \textbf{79.5}±1.1 & 69.2±1.2 & \textbf{86.2}±0.2 & \textbf{86.5}±0.2 & \textbf{80.3}±0.0 & \textbf{+0.7}\\ \midrule
 & \multicolumn{5}{c}{With $\mathcal{O}$ (\textbf{GReg1})} \\
\textbf{GAim2 + GReg1} (MIRO, SIMPLE) & 83.2±0.6 & 72.6±1.1 & 89.9±0.5 & 90.2±0.1 & 84.0±0.2 & 0.0\\
\textbf{GAim2 + GReg1 +iAim1} & 83.4±0.5 & 73.1±0.8 & 89.7±0.4 & 90.1±0.3 & 84.1±0.2 & +0.1\\
\textbf{GAim2 + GReg1 + GAim1} & 83.7±0.3 & 74.0±0.6 & 90.1±0.3 & 90.3±0.2 & 84.5±0.2 & +0.4\\
\textbf{GAim2 + GReg1 + iReg2} & 82.9±0.5 & 72.5±0.3 & \textbf{90.3}±0.3 & 90.0±0.3 & 83.9±0.1 & -0.1\\
\textbf{GAim2 + GReg1 + GReg2} & 83.4±0.2 & 72.3±0.2 & 90.1±0.3 & 90.1±0.3 & 84.0±0.2 & +0.0\\ 
\rowcolor{mygray} 
\begin{tabular}[c]{@{}l@{}}\textbf{GAim2 + GReg1} \\ \textbf{+ GAim1 + GReg2 (\blue{\textbf{GMDG}})} \end{tabular}& \textbf{84.1}±0.2 & \textbf{74.3}±0.9 & 89.9±0.4 & \textbf{90.6}±0.1 & \textbf{84.7}±0.2 & \textbf{+0.7}\\ \bottomrule
\end{tabular}%
}
\vspace{-5pt}
\caption{Ablation studies: Results of using different combinations of terms on HomeOffice. Imp. denotes Improvement that gained form \textbf{GAim2} and \textbf{GAim2 + GReg1}, respectively.} 
\vspace{-5pt}
\label{tab:abl_res}
\end{table}

{1). Previous methods that partially utilize our proposed objectives often yield suboptimal results.} 
Note that \textbf{iAim1} is the unconditional version of \textbf{GAim1}.
By eliminating other factors, it can be seen that employing our proposed full objectives offers the most significant improvements, 
while previous objectives may lead to inferior results. 

{2). The effectiveness of using conditions.} By conducting uniform implementation and testing, it can be observed that the use of conditions yields superior results compared to the unconditional approach. This observation aligns with Eq.~\ref{eq:risk_compare_cond},  suggesting that minimizing the gap between conditional features across domains leads to improved generalization. The disparity in performance between CDANN and DANN  might be attributed to differences in their implementation details. 

{3). Learning invariance is crucial, regardless of whether integrating prior knowledge.} 
Evidently, learning invariance facilitates improvement whether prior is applied or not, as validated in the PUB derivation in Eq.~\ref{eq:PUB}. 
This contradicts MIRO's argument that achieving similar representations to a prior can replace the need for learning invariance.    

{4). Impacts of using prior.} 
The significant improvement owes to the use of a pre-trained oracle model ($\mathcal{O}$) preserving correlations between $\rmX$ and $\rmY$ - a concept validated by MIRO and SIMPLE.
However, utilizing our full set of objectives can further enhance this improvement by an additional $0.7\%$. Notably, the invalid causality may not work when using prior knowledge, while the invariance across domains is not permitted.
We hypothesize that such invalid causality is inherently eliminated within a `good' feature space obtained by $\mathcal{O}$, but may be reintroduced when we minimize the domain gap with $\mathcal{O}$.
Thus, using the full objective can synergistically produce optimal results. 
5).
Constraining only the covariance shifts of features across domains (\textbf{iReg2}) does not guarantee better results when prior knowledge is available.
We find that using the objectives of CORAL performs better than DANN, CDANN, and CIDG. 
The results suggest that considering the covariance shifts of features does lead to improvements, which we hypothesize are primarily driven by $H(P(\phi(\rmX)))$.
However, when a large pre-trained oracle model ($\mathcal{O}$) is provided, the performance actually degrades.
This implies that the use of $\mathcal{O}$ implicitly minimizes the covariance shifts of features across domains.
Under this scenario, the unexpected effect of $-H( P ({\phi(\rmX|\mathcal{D})}))$ hinders improvement, while the benefits brought by $H(P(\phi(\rmX)))$ are diminished by the use of prior knowledge. 
In contrast, \textbf{GReg2} continues to yield improvements. This suggests that GMDG is more versatile and suitable for various situations.

\section{Conclusion}
In this paper, we propose a general objective, namely GMDG, by relaxing the static distribution assumption of $\rmY$ through a learnable mapping $\psi$. 
GMDG is applicable to diverse mDG tasks, including regression, segmentation, and classification.  Empirically, we design a suite of losses to achieve the overall GMDG, adaptable across various frameworks. 
Extensive experiments validate the viability of our objective across applications where previous objectives may yield suboptimal results compared to ours. Both theoretical analyses and empirical results demonstrate the synergistic effect of distinct terms in the proposed objective.  Simplistically, we assume equal domain weights whilst minimizing GJSD, presenting the future scope for dealing with imbalance situations triggering unequal domain weights.

\section*{Acknowledgments}
The work was partially supported by the following: National Natural Science Foundation of China under No. 92370119, No. 62376113, and No. 62206225; Jiangsu Science and Technology Program (Natural Science Foundation of Jiangsu Province) under No.  BE2020006-4;
Natural Science Foundation of the Jiangsu Higher Education Institutions of China under No. 22KJB520039.

{
    \small
    \bibliographystyle{ieeenat_fullname}
    \bibliography{main}
}



\clearpage
\setcounter{page}{1}
\maketitlesupplementary


\section{More mathematical details of our method}
\label{app:our_obj}
\subsection{Derivation details of PUB}

\textbf{Details for Eq.~\ref{eq:bound1}.}

We denote $P_{mix} \triangleq \sum_n w_n P({\phi(\rmX_n), \psi(\rmY_n)}) $.  Therefore for GJSD, we have:
\begin{equation}
    \begin{split}
             & GJSD(\{P ({\phi(\rmX_n), \psi(\rmY_n)})\}_{n=1}^{N}) 
     \\ 
     = &\sum\nolimits_n w_n KL (P(\phi(\rmX_n) ,  \psi(\rmY_n))\Vert P_{mix}) 
      \\
     = &\sum\nolimits_n w_n [H_c( P(\phi(\rmX_n), \psi(\rmY_n)), P_{mix})  \\ 
    &- H(P(\phi(\rmX_n), \psi(\rmY_n)))] 
      \\
     = &\sum\nolimits_n w_n H_c( P(\phi(\rmX_n), \psi(\rmY_n)), P_{mix})  \\ 
    &- \sum\nolimits_n w_n H(P(\phi(\rmX_n), \psi(\rmY_n)))
      \\
     = &\sum\nolimits_n w_n \int_{\phi(\rmX_n), \psi(\rmY_n)} - P(x,y) \ln P_{mix}(x,y) d(x,y)   \\ 
    &- \sum\nolimits_n w_n H(P(\phi(\rmX_n), \psi(\rmY_n)))  
      \\
     = &\int_{\phi(\rmX_n), \psi(\rmY_n)} - \sum\nolimits_n w_n P(x, y) \ln P_{mix}(x,y) d(x,y) \\ 
    & - \sum\nolimits_n w_n H(P(\phi(\rmX_n), \psi(\rmY_n))) 
      \\
      = &\int_{\phi(\rmX_n), \psi(\rmY_n)} - P_{mix}(x,y) \ln P_{mix}(x,y) d(x,y)  \\ 
    &- \sum\nolimits_n w_n H(P(\phi(\rmX_n), \psi(\rmY_n))) 
      \\
     = & H(P_{mix}) - \sum\nolimits_n w_n H(P(\phi(\rmX_n), \psi(\rmY_n))).
    \end{split}
\end{equation}

Therefore, the minimization of GJSD can be written as follows:
\begin{equation}
    \resizebox{\hsize}{!}{$\begin{split}
    & \min_{\phi, \psi} GJSD(\{P ({\phi(\rmX_n), \psi(\rmY_n)})\}_{n=1}^{N}) 
     \\
    \equiv &\min_{\phi, \psi} H(\mathbb{E}[ P ({\phi(\rmX_n), \psi(\rmY_n)})]) - \mathbb{E}[ H (P ({\phi(\rmX_n), \psi(\rmY_n)}))] \\
     \equiv &\min_{\phi, \psi}  H( P ({\phi(\rmX), \psi(\rmY)} \mid \mathcal{D})) - \mathbb{E} [H (P ({\phi(\rmX_n), \psi(\rmY_n)}))]. 
    \end{split}$}
\end{equation}


\textbf{Details for Eq.~\ref{eq:bound2}.} Taking into account $\mathcal{O}$,  similar to \citep{cho2022cooperative}, we have the upper bound for GJSD as:
\begin{equation}
    \begin{split}
         & GJSD(\{P ({\phi(\rmX_n), \psi(\rmY_n)})\}_{n=1}^{N}) 
     \\ 
     = & H_c(P_{mix}, \mathcal{O}) -   H_c(P_{mix}, \mathcal{O}) + H(P_{mix})   \\ 
    & -\sum_n w_n H(P(\phi(\rmX_n), \psi(\rmY_n))) 
      \\
     = & H_c(P_{mix}, \mathcal{O}) - \KL( P_{mix} \Vert \mathcal{O})  \\ 
    & - \sum_n w_n H(P(\phi(\rmX_n), \psi(\rmY_n))) 
      \\
     \leq & H_c(P_{mix}, \mathcal{O}) - \sum_n w_n H(P(\phi(\rmX_n), \psi(\rmY_n))). 
    \end{split}
\end{equation}
For the standard situation where $w_1 = w_2 = ... =  w_n = 1/N$, we further have:
\begin{equation}
    \begin{split}
        & GJSD(\{P ({\phi(\rmX_n), \psi(\rmY_n)})\}_{n=1}^{N}) 
     \\
     \leq & H_c(P_{mix}, \mathcal{O}) - \sum_n w_n H(P(\phi(\rmX_n), \psi(\rmY_n))) 
      \\
     = & H_c(\mathbb{E}[ P ({\phi(\rmX_n), \psi(\rmY_n)})],\mathcal{O}) \\ 
     &- \mathbb{E}[ H(P(\phi(\rmX_n), \psi(\rmY_n)))].
    \end{split}
\end{equation}

\textbf{Details for Eq.~\ref{eq:PUB}.} The above bound can be further reformed as:
\begin{equation}
    \begin{split}
         & H_c(\mathbb{E}[ P ({\phi(\rmX_n), \psi(\rmY_n)})],\mathcal{O})- a 
      \\
       =  & H(\mathbb{E}  [P({\phi(\rmX_n), \psi(\rmY_n)})])  \\ 
    & + \KL(\mathbb{E}[P({\phi(\rmX_n), \psi(\rmY_n)})] \Vert \mathcal{O})  - a,
         \\
      =  & H(\mathbb{E}  [P({\phi(\rmX_n), \psi(\rmY_n)})])  \\ 
    & + \KL(P({\phi(\rmX), \psi(\rmY)}) \Vert \mathcal{O})  - a.
    \end{split}
\end{equation}

\textbf{Derivation details of Eq.~\ref{eq:GAim1_GReg2}.}
\begin{equation}
    \begin{split}
     \text{\textbf{GAim1}} = &H( P ({\phi(\rmX), \psi(\rmY)} \mid \mathcal{D})) \\
     = & H(P (\phi(\rmX), \psi(\rmY), \mathcal{D})) -H(P(\mathcal{D}))\\ 
     = & H(P (\phi(\rmX), \mathcal{D})) - H(P(\mathcal{D})) \\ & +  H(P(\psi(\rmY)\mid \phi(\rmX), \mathcal{D})) \\ 
    = & H(P (\phi(\rmX )\mid \mathcal{D})) +  H(P(\psi(\rmY)\mid \phi(\rmX), \mathcal{D})) \\ 
    = & H(P (\psi(\rmY)\mid \mathcal{D})) +  H(P(\phi(\rmX) \mid \psi(\rmY),  \mathcal{D})). 
    \end{split}
\end{equation}

\textbf{Derivation details of Eq.~\ref{eq:why_greg2}.}
Due to Eq.~\ref{eq:GAim1_GReg2}, we want to maintain $\phi(\rmX) \to \psi(\rmY)$ but suppressing   $\psi(\rmY) \to \phi(\rmX)$. Thus we want to $ \max_{\phi,\psi} H(P(\phi(\rmX) | \psi(\rmY)), \mathcal{D})$ while $\min _{\phi,\psi}H(P( \psi(\rmY) | \phi(\rmX) ), \mathcal{D})$. which problem can be simplified as:
\begin{equation}
    \begin{split}
     &\min_{\phi,\psi} H(P( \psi(\rmY) | \phi(\rmX) , \mathcal{D} ))-H(P(\phi(\rmX) | \psi(\rmY) , \mathcal{D})) \\
     =& \min_{\phi,\psi} H(\phi(\rmX),\psi(\rmY)|\mathcal{D})) -  I(\phi(\rmX),\psi(\rmY)|\mathcal{D})) - \\
     & H(P(\phi(\rmX) | \psi(\rmY) , \mathcal{D})) + H(P(\phi(\rmX) | \psi(\rmY) , \mathcal{D})) \\
      =& \min_{\phi,\psi} H(\phi(\rmX),\psi(\rmY)|\mathcal{D})) -  I(\phi(\rmX),\psi(\rmY)|\mathcal{D})),
    \end{split}
\end{equation}
where the first term is already in \textbf{GAim2}, thus \textbf{GReg2} should deal with the second term, which is:
\begin{equation}
    \begin{split}
     &\min_{\phi,\psi} I(\phi(\rmX),\psi(\rmY)|\mathcal{D})) = \\ & \min_{\phi,\psi} H (P(\phi(\rmX))|\mathcal{D}) - H(P(\phi(\rmX))|P(\psi(\rmY)),\mathcal{D}).
    \end{split}
\end{equation}
Also, due to the effect of $\mathcal{D}$ being alleviated through the mappings, the above equation is approximated as  
\begin{equation}
    \begin{split} 
    \min_{\phi,\psi} H (P(\phi(\rmX))) - H(P(\phi(\rmX))|P(\psi(\rmY))),
    \end{split}
\end{equation}
which is \textbf{GReg2}.

\textbf{Derivation details of Eq.~\ref{eq:cfs_detial}.}
For a Gaussian distribution $\mathcal{N}(x; \mu, \Sigma)$ with $D$ dimension, its entropy is:
\begin{equation}
    \begin{split}
        H(x) = & - \int p(x) \ln{p(x)} dx 
    \\ 
    = & - \int p(x) [\ln{((2\pi)^{-\frac{D}{2}}|\Sigma|^{-\frac{1}{2}})}  
    \\&  - \frac{1}{2}(x-\mu)^\top \Sigma^{-1} (x-\mu) ] dx
     \\ 
    = & \ln{((2\pi)^{\frac{D}{2}}|\Sigma|^{-\frac{1}{2}}) }  
    \\&  + \frac{1}{2} \int p(x)[(x-\mu)^\top \Sigma^{-1} (x-\mu)] dx
     \\ 
    = & \ln{((2\pi)^{\frac{D}{2}}|\Sigma|^{-\frac{1}{2}}) } + \frac{1}{2} \int p(y) \times y^{\top} dy
     \\ 
     = & \ln{((2\pi)^{\frac{D}{2}}|\Sigma|^{-\frac{1}{2}}) } + \frac{1}{2} \sum_{d=1}^D \mathbb{E}[y_d^2]
     \\ 
     = & \ln{((2\pi)^{\frac{D}{2}}|\Sigma|^{-\frac{1}{2}}) } + \frac{D}{2}
      \\
    = & \frac{D}{2}(1 + \ln{(2\pi) }) + \frac{1}{2} \ln{|\Sigma|}.
    \end{split}
\end{equation}

Then Eq.~\ref{eq:cfs_detial} equals: 
\begin{equation}
\label{eq:cfs_detial_proof}
    \begin{split}
        & H( \mathcal{N}(\phi(\rmX); \mu_\rmX, \Sigma_{\rmX\rmX}))  
    \\& - H( \mathcal{N}(\phi(\rmX)\mid \psi(\rmY); \mu_{\rmX|\rmY}, \Sigma_{\rmX\rmX|\rmY}))  
    \\
    =&\frac{D}{2} (1 + \ln(2\pi)) + \frac{1}{2} \ln| \Sigma_{\rmX\rmX} |  
    \\&  -  \frac{D}{2} (1 + \ln(2\pi)) - \frac{1}{2} \ln| \Sigma_{\rmX\rmX|\rmY} |  
     \\
     = & \frac{1}{2} \ln(|\Sigma_{\rmX\rmX}|) - \ln(|\Sigma_{\rmX\rmX|\rmY}|), 
    \\
     = & \frac{1}{2} \ln(\frac{|\Sigma_{\rmX\rmX}|}{|\Sigma_{\rmX\rmX|\rmY}|}). 
    \end{split}
\end{equation}

\textbf{Empirical risk.}
The empirical risk introduced by the whole model $\theta$ \textit{w.r.t} $\rmX, \rmY$ is determined by a convex loss function $L(\theta)$. Following ~\citep{perlaza2022empirical},
the empirical risk considering $\mathcal{O}$ is:
 \begin{align}
     R(\theta) = & \int L(\theta)dP(\theta) +  H(\mathbb{E}[ P ({\phi(\rmX_n), \psi(\rmY_n)})]) \nonumber 
    \\&  + \KL( \mathbb{E}[ P ({\phi(\rmX_n), \psi(\rmY_n)})] \Vert \; \mathcal{O}) \\ & - H(P(\phi(\rmX)  \mid  \psi(\rmY))) 
        + H(P(\phi(\rmX))) . \nonumber
\end{align}

\textbf{Proof of \textbf{Using $\psi$ \textit{v.s.} not using $\psi$}.} 
Using \textit{ Jensen’s inequality}, due to $\rmY, \psi(\rmY)$ contains the same amount of useful information as $\rmY$, we have:
\begin{align}
    H(\rmY) \ge H(\psi(\rmY)).
\end{align}

Therefore, we have
\begin{equation}
    \begin{split}
    &H(\mathbb{E}[ P ({\phi(\rmX_n), \rmY_n})]) \\
     =  &H(\mathbb{E}[ P (\phi(\rmX_n))]) + H(\mathbb{E}[ P ({\rmY_n \mid \phi(\rmX_n) })])  \\
     \ge & H(\mathbb{E}[ P (\phi(\rmX_n))]) + H(\mathbb{E}[ P ({\psi(\rmY_n )\mid \phi(\rmX_n) })])   \\
     = &H(\mathbb{E}[ P ({\phi(\rmX_n), \psi(\rmY_n)})]) .  
    \end{split}
\end{equation}
Therefore, for the risk of $\theta^{n\psi}$:
 \begin{align}
     \sup R({\theta}^{n\psi}) =  \sup & \min_{\phi} [\int L(\theta)dP(\theta) \nonumber 
    \\& +  H(\mathbb{E}[ P ({\phi(\rmX_n), \rmY_n})])] + b,
 \end{align}
and the risk of $\theta^{n\psi}$:
\begin{align}
     \sup R({\theta}^{\psi}) =  \sup & \min_{\phi} [\int L(\theta)dP(\theta) \nonumber 
    \\& +  H(\mathbb{E}[ P ({\phi(\rmX_n), \psi(\rmY_n)})])] + b,
 \end{align}
 where $b \triangleq \KL( \mathbb{E}[ P ({\phi(\rmX_n), \psi(\rmY_n)})] \Vert \; \mathcal{O})   - H(P(\phi(\rmX)  \mid  \psi(\rmY))) + H(P(\phi(\rmX))) , $  we have:
 \begin{align}
     \sup R({\theta}^{n\psi}) \ge \sup R({\theta}^{\psi}) .
 \end{align}

\textbf{Proof of incorporating conditions leads to lower generalization risk on learning invariant representations.}
For the risks of the model having parameters $\theta^{c}$ trained with using conditions, we have:
 \begin{align}
     \sup R({\theta}^{c}) =  \sup & \min_{\phi} [\int L(\theta)dP(\theta) \nonumber 
    \\& +  H(\mathbb{E}[ P ({\phi(\rmX_n), \psi(\rmY_n)})])] + b,
 \end{align}
 where $b \triangleq \KL( \mathbb{E}[ P ({\phi(\rmX_n), \psi(\rmY_n)})] \Vert \; \mathcal{O})   - H(P(\phi(\rmX)  \mid  \psi(\rmY))) + H(P(\phi(\rmX))) . \nonumber$ For $R({\theta}^{nc})$ that trained without using conditions, it has:
  \begin{align}
     \sup R({\theta}^{nc}) = \sup & \min_{\phi,\psi} [ \; \int L(\theta)dP(\theta)  +  H(\mathbb{E}[ P ({\phi(\rmX_n)})])  \;  ] \nonumber 
    \\&  + H(\mathbb{E}[ P ({\phi(\rmX_n), \psi(\rmY_n)})])  \nonumber \\ 
     & -  H(\mathbb{E}[ P ({\phi(\rmX_n)})]) + b   \\
      = \sup & \min_{\phi,\psi} [\;\int L(\theta)dP(\theta)  +  H(\mathbb{E}[ P ({\phi(\rmX_n)})]) \;] \nonumber 
    \\& + H(\mathbb{E}[ P ({ \psi(\rmY_n) \mid \phi(\rmX_n)})]) +b. \nonumber 
 \end{align}
Due to the inequality:
\begin{align}
    &\sup  \min_{\phi} [ \; H(\mathbb{E}[ P ({\phi(\rmX_n), \psi(\rmY_n)})]) \;] 
    \\ = &\sup  \min_{\phi} [ \; H(\mathbb{E}[ P ({\phi(\rmX_n)})]) \nonumber 
    \\& \underbrace{ + H(\mathbb{E}[ P ({ \psi(\rmY_n) \mid \phi(\rmX_n)})])}_{\text{Minimized.}} \;] 
    \\\leq & \sup  \min_{\phi,\psi} [ \;H(\mathbb{E}[ P ({\phi(\rmX_n)})]) \;] \nonumber 
    \\& \underbrace{+ H(\mathbb{E}[ P ({ \psi(\rmY_n) \mid \phi(\rmX_n)})])}_{\text{Remains.}},
\end{align}
we have
\begin{align}
     \sup R({\theta}^{nc}) \ge \sup R({\theta}^{c}) .
\end{align}


\section{Objective derivation details of many previous methods.}
\label{app:pre_obj}

This section shows how we uniformly simplify the objectives of previous methods. 



\textbf{ERM~\citep{gulrajani2020search}: The basic method.} 
The basic method does not focus on minimizing GJSD. Therefore, there are no terms for \textbf{Aim~1}. For \textbf{Aim~2} it directly minimize $H(P ({\phi(\rmX), \rmY}))$.

\textbf{DANN~\citep{ganin2016domain}: Minimize feature divergences of source domains.} DANN~\citep{ganin2016domain} minimizes feature divergences of source domains adverbially without considering conditions. Therefore its empirical objective for \textbf{Aim~1} is 
\begin{align}
    &\min_{\phi} H(\mathbb{E}[ P ({\phi(\rmX_n)})]) -a
\end{align}
For \textbf{Aim~2} it directly minimizes $H(P ({\phi(\rmX), \rmY}))$.

\textbf{CORAL~\cite{sun2016deep}: Minimize the distance between the second-order statistics of source domains.} Since CORAL~\cite{sun2016deep} only minimizes the second-order distance between souce feature distributions, its objective
can be summarized as:
\begin{align}
   \min_{\phi} 
    H(P ({ \phi(\rmX) , \rmY}))+ H(P({\phi(\rmX)})) -H(\mathbb{E}[ P ({\phi(\rmX_n)})]) .
\end{align}
By grouping it, CORAL~\citep{sun2016deep} has $ -H(\mathbb{E}[ P ({\phi(\rmX_n)})])$ for \textbf{Aim~1} and $H(P ({ \phi(\rmX) , \rmY}))+ H(P({\phi(\rmX)}))$ for \textbf{Aim~2}.

\textbf{CIDG~\citep{li2018domain}: Minimizing the conditioned domain gap.} CIDG~\citep{li2018domain} tries to learn conditional domain invariant features:  
\begin{align}
\label{eq:CIDG} 
    \min_{\phi} H(\mathbb{E}[ P ({\phi(\rmX_n), \rmY_n})]).
\end{align}
For \textbf{Aim~2} it directly minimizes $H(P ({\phi(\rmX), \rmY}))$.

\textbf{MDA~\citep{hu2020domain}: Minimizing domain gap compared to the decision gap.} 
Some previous works, such as MDA~\citep{hu2020domain}, follow the hypothesis that the generalization is guaranteed while the decision gap is larger than the domain gap. Therefore, instead of directly minimizing the domain gap, MDA minimizes the ratio between the domain gap and the decision gap. The overall objective of MDA can be summarized as:
\begin{equation}
    \begin{split}
    \label{eq:MDA}
    &\min_{\phi}  H(P ({\phi(\rmX), \rmY})) + H(P (\phi(\rmX)))  
    \\& +  ({H(\mathbb{E}[ P ({\phi(\rmX_n), \rmY_n})]) - \underbrace{\mathbb{E}[H (P ({\phi(\rmX_n),\rmY_n}))]}_{constant}})  \\
    & - ({H(\mathbb{E}[ P({\phi(\rmX_n)\mid \rmY_n})]) - \underbrace{ \mathbb{E}[H (P ({\phi(\rmX)\mid \rmY}))]}_{constant}})  
    \\& + \underbrace{\mathbb{E}[H(P ({\phi(\rmX), \rmY}))]}_{constant} .
    \end{split}
\end{equation}


    Since  the entropy is non-negative and the constants can be omitted, Eq.~\ref{eq:MDA} is equivalent to:
\begin{equation}
    \begin{split}
    \label{eq:MDA2}
    &\min_{\phi} H(P ({\phi(\rmX)\mid \rmY})) + H(P (\phi(\rmX))) 
    \\
    & + H(\mathbb{E}[ P ({\phi(\rmX_n), \rmY_n})])-  H(\mathbb{E}[ P({\phi(\rmX)\mid \rmY})])) + a.  
    \end{split}
\end{equation}
By grouping Eq.~\ref{eq:MDA2}, we have that for \textbf{Aim~1} it minimizes $\min_{\phi} H(\mathbb{E}[ P ({\phi(\rmX_n), \rmY_n})])$, and for \textbf{Aim~2} it minimizes $H(P ({\phi(\rmX)\mid\rmY})) -  H(\mathbb{E}[ P({\phi(\rmX) \mid \rmY})])) +  H(P (\phi(\rmX)))$.

\textbf{MIRO~\citep{cha2022miro}, SIMPLE~\citep{li2022simple}: Using pre-trained models as $ \mathcal{O}$.} 
One feasible way to obtain $\mathcal{O}$ is adopting pre-trained oracle models such as MIRO~\citep{cha2022miro} and SIMPLE~\citep{li2022simple}. Note that the pre-trained models are exposed to additional data besides those provided. Therefore, for \textbf{Aim~1}: they have: 
\begin{align}
\label{eq:orcale_bound}
    \min_{\phi} \KL( P (\phi(X)|Y) \Vert\mathcal{O})-  \underbrace{\mathbb{E}[ H (P ({\phi(\rmX_n),\rmY_n}))]}_{constant}.
\end{align}
Differently, MIRO only uses one pre-trained model, as its $\mathcal{O} \triangleq  \mathcal{O}^1$; meanwhile, SIMPLE combines $K$ pre-trained models as the oracle model: $\mathcal{O} \triangleq \sum_{k=1}^{K} v_k  \mathcal{O}^k$ where $v$ is the weight vector.  
For \textbf{Aim~2} it directly minimizes $H(P ({\phi(\rmX), \rmY}))$.

\textbf{RobustNet~\citep{choi2021robustnet}.} RobustNet employs the instance selective whitening loss, which disentangles domain-specific and domain-invariant properties from higher-order statistics of the feature representation and selectively suppresses domain-specific ones. Therefore, it implicitly whitens the $\rmY$-irrelevant features in $\rmX$. Thus, its objective can be simplified as:
\begin{equation}
    \begin{split}
         &\min_{\phi,{\psi}}H(P ({ \phi(\rmX) , {\psi(\rmY)}}))  - H(P(\phi(\rmX)  \mid  \psi(\rmY))) 
    \\&   + H(P(\phi(\rmX))).
    \end{split}
\end{equation}

\section{Aligning notations between paper and supplementary materials}
\subsection{More details about Table~\ref{tab:summary}}
To better understand, we simplify some notations in Table~\ref{tab:summary}. We present the simplified notations and their corresponding origins in Table~\ref{tab:summary2}.

\begin{table*}[t]
\begin{center}
\begin{tabular}{lll}
\toprule
 & \multicolumn{2}{c}{ \bf Learning domain invariant representations}\\ 
 & \multicolumn{1}{c}{\bf Aim1: Learning domain invariance} & \multicolumn{1}{c}{\bf Reg1: Integrating prior} \\ \midrule
\specialrule{0em}{3pt}{3pt}
DANN & $\min_{\phi} H(  P (\phi(\rmX) \mid \mathcal{D}))$ & None \\
  & \blue{$\min_{\phi} H( \mathbb{E}[ P (\phi(\rmX_{n})) ]$} &  \\ \specialrule{0em}{3pt}{3pt}
CDANN, CIDG, MDA & $\min_{\phi} H(  P ({\phi(\rmX), \rmY} \mid \mathcal{D}))$ & None \\
  & \blue{$\min_{\phi} H( \mathbb{E}[ P (\phi(\rmX_{n}), \psi( \rmY_n)) ]$} &  \\ \specialrule{0em}{3pt}{3pt}
\textbf{Ours} & {$\min_{\phi, \psi} H( P ({\phi(\rmX), \psi(\rmY)} \mid \mathcal{D}))$} & {${\min_{\phi, \psi}\KL(P({\phi(\rmX), \psi(\rmY)})\Vert \mathcal{O})}$} \\
 & \blue{$\min_{\phi, \psi} H(  \mathbb{E}[P ({\phi(\rmX_N), \psi(\rmY_n)}])$} \\ \specialrule{0em}{3pt}{3pt}
\bottomrule \toprule
 & \multicolumn{2}{c}{ \bf Maximizing A Posterior between representations and targets}\\ 
 & \multicolumn{1}{c}{ \bf Aim2: Maximizing A Posterior (MAP)}   & \multicolumn{1}{c}{ \bf Reg2: Suppressing invalid causality}\\ \midrule \specialrule{0em}{3pt}{3pt}
CORAL & {$\min_{\phi}H(P ({\rmY \mid  \phi(\rmX)}))$} & {$\min_{\phi}-H( P ({\phi(\rmX\mid \mathcal{D} )})) + H(P({\phi(\rmX)}))$} \\ 
  &   & \blue{$\min_{\phi}-H( \mathbb{E}[P ({\phi(\rmX_n)})]) + H(P({\phi(\rmX)}))$} \\ 
  \specialrule{0em}{3pt}{3pt}
\bottomrule
\end{tabular}%
\caption{Supplemental notations for Table~\ref{tab:summary}. Refined notations and their original formulations are reported. The original formulations are highlighted as \blue{blue}. }
\label{tab:summary2}
\end{center}
\end{table*}

\subsection{More details about Table~\ref{tab:note}}
For simplification, we uniformly simplified the formulation of terms from their derivation. The simplified form in Table~\ref{tab:note} and its original form can be seen in  Table~\ref{tab:note_align}. Note that the \textbf{iAim1} is from CDANN~\cite{li2018deep}, CIDG~\citep{li2018domain}, MDA~\citep{hu2020domain} and \textbf{iReg2} is from CORAL~\citep{sun2016deep}.

\begin{table}[t]
    \centering
\begin{tabular}{llll}
\toprule
\textbf{GAim2} & $H(P ({\psi(\rmY) \mid  \phi(\rmX)})) + H(P ({\rmY \mid  \psi(\rmY)}))$ &  \\ 
\textbf{GReg1} & $\KL(P ({\phi(\rmX), \rmY} \mid \mathcal{D}) \Vert \mathcal{O}))$  \\ 
\textbf{iAim1} & $H( P ({\phi(\rmX)} \mid \mathcal{D}))$ & \\ 
\textbf{GAim1} & $ H(  P ({\phi(\rmX), \rmY}) \mid \mathcal{D})$  \\
\textbf{iReg2} & $ -H( P ({\phi(\rmX)} , \mathcal{D}) + H(P({\phi(\rmX)}))$ & \\ 
\textbf{GReg2} & ${ - H(P(\phi(\rmX)  \mid  \psi(\rmY))) 
+ H(P(\phi(\rmX)))}$ \\  
\bottomrule
\toprule
\textbf{GAim2} & $H(P ({\psi(\rmY) \mid  \phi(\rmX)})) + H(P ({\rmY \mid  \psi(\rmY)}))$ & \\ 
\textbf{GReg1} & $ \KL(\mathbb{E}[ P ({\phi(\rmX_n), \rmY_n})] \Vert \mathcal{O}))$  \\ 
\textbf{iAim1} &  $H(\mathbb{E}[ P ({\phi(\rmX_n)})])$ & \\ 
\textbf{GAim1} &  $ H(\mathbb{E}[ P ({\phi(\rmX_n), \rmY_n})])$  \\
 \textbf{iReg2} & $ -H(\mathbb{E}[ P ({\phi(\rmX_n)})]) + H(P({\phi(\rmX)}))$ & \\ 
 \textbf{GReg2} & ${ - H(P(\phi(\rmX)  \mid  \psi(\rmY))) 
+ H(P(\phi(\rmX)))}$ \\  
\bottomrule
\end{tabular}%
\vspace{-5pt}
\caption{Notations for terms in the paper (above) and its derived formulation (below) in the appendix.}
\label{tab:note_align}
\end{table}

\section{Experimental details and parameters}
\label{app:para}
\textbf{We have conducted  $248$ experiments in total,} including 
$12$ Toy experiments (training $3$ objective settings on $4$ domain settings), 
$20$ Regression experiments in Monocular depth estimation (training $5$ objective settings on $4$ domain settings), 
$9$ Segmentation experiments (training $3$ objective settings on $1$ domain settings and verifying on $3$ domain settings), 
$63$ Classification experiments (training $1$ objective settings on $5$ datasets that has $4,4,4,4,5$ domain settings for $3$ trails), 
and $144$ ablation study experiments (training $12$ objective settings on $1$ dataset that has $4$ domain settings for $3$ trails).
We believe that the consistent improvements yielded by GMDG in these experiments validate the superiority of our GMDG.

Experimental details of these experiments can be found in the following.
Note that we set $v_{A2} = 1$  for all experiments. 

\subsection{Toy experiments: Synthetic regression experimental details.}
We explore the efficacy of $\psi$ by using toy regression experiments with synthetic data.

\textbf{Datasets.} The latent features in all three domains are added some distributional shifts and used as the first group in the raw features (denoted as $x^1_n, y^1_n$ where $n\in{1,2,3}$ represent which domain it belongs to). Then, some domain-conditioned transformations are applied to shifted features, or some pure random noises are used as the second group in the raw features (denoted as $x^2_n, y^2_n$). 
Therefore the constructed 
$X_{n\in\{1,2,3\}}= [x^1_n, x^2_n], Y_{n\in\{1,2,3\}} = [y^1_n,y^2_n]$  both contain features that dependents on $D$. 
Details of each synthetic data are exhibited in Table~\ref{tab:syn}. We generate $10000$ samples for training and $100$ samples for validation and testing sets.

\textbf{Parameter settings.} All experiments are conducted with $v_{A1}, v_{R1}, v_{R2}= 0.1$.

\textbf{Experimental settings.} For $\phi,\psi$, we use three-layer MLP and one linear layer for regression prediction and Mean Squared Error (MSE) as the loss. We use the best model on the validation dataset for testing. 

\textbf{Metric.} We use the MSE between the predictions and the $Y$ of the testing set as the evaluation metric.

\begin{table*}[t]
\centering
\begin{tabular}{lllll}
\toprule
$hx$ & $hx \sim \mathcal{N}(hx; 0, 1)$ \\
$hy$ & $hy= hx$ \\  \midrule 
Data 1 & Without distribution shift & With affine transformations \\
$X_1$ & $x_1^1 =hx$ & $x_1^2 = x_1^1 +\epsilon\sim \mathcal{N}(\epsilon; 0, 0.3)$ &  \\
$Y_1$ & $y_1^1 =hy$ & $y_1^2 = y_1^1 +\epsilon\sim \mathcal{N}(\epsilon; 0, 0.3)$ &  \\ \hline
$X_2$ & $x_2^1 =hx$ & $x_2^2 = 4 \times x_2^1 + \epsilon \sim \mathcal{N}(\epsilon; 0.5, 0.3)$ &  \\
$Y_2$ & $y_2^1 =hy$ & $y_2^2 = 4 \times y_2^1 + 0.3 $ &  \\ \hline
$X_3$ & $x_3^1 =hx$ & $x_3^2 = 2 \times x_3^1 + \epsilon \sim \mathcal{N}(\epsilon; -0.3, 0.2)$ &  \\
$Y_3$ & $y_3^1 =hy$ & $y_3^2 = 0.5 \times y_3^1 - 0.2$ &  \\
\midrule 
Data 2 & With distribution shift &  With affine transformations \\
$X_1$ & $x_1^1 =hx$ & $x_1^2 = x_1^1 +\epsilon\sim \mathcal{N}(\epsilon; 0, 0.3)$ &  \\
$Y_1$ & $y_1^1 =hy$ & $y_1^2 = y_1^1+\epsilon\sim \mathcal{N}(\epsilon; 0, 0.3)$ &  \\ \hline
$X_2$ & $x_2^1 =hx+\epsilon\sim \mathcal{N}(\epsilon; -0.1, 0.1)$ & $x_2^2 = 4 \times x_2^1 + \epsilon \sim \mathcal{N}(\epsilon; 0.3, 0.3)$ &  \\
$Y_2$ & $y_2^1 =hy+\epsilon\sim \mathcal{N}(\epsilon; 0.2, 0.1)$  & $y_2^2 = 8 \times y_2^1 - 0.3 $ &  \\ \hline
$X_3$ & $x_3^1 =hx+\epsilon\sim \mathcal{N}(\epsilon; 0.4, 0.2)$  & $x_3^2 = -1 \times x_3^1 +\epsilon \sim \mathcal{N}(\epsilon; -0.3, 0.2)$ &  \\
$Y_3$ & $y_3^1 =hy+\epsilon\sim \mathcal{N}(\epsilon; -0.4, 0.2)$ & $y_3^2 = \epsilon \sim \mathcal{N}(\epsilon; 0, 0.2)$ & 
\\
\midrule 
Data 3 & Without distribution shift & With squared, cubed transformations or noises\\
$X_1$ & $x_1^1 =hx$ & $x_1^2 = x_1^1 +\epsilon\sim \mathcal{N}(\epsilon; 0, 0.3)$ &  \\
$Y_1$ & $y_1^1 =hy$ & $y_1^2 = y_1^1+\epsilon\sim \mathcal{N}(\epsilon; 0, 0.3)$ &  \\ \hline
$X_2$ & $x_2^1 =hx$ & $x_2^2 = 4 \times x_2^1 ** 3 + \epsilon \sim \mathcal{N}(\epsilon; 0.5, 0.3)$ &  \\
$Y_2$ & $y_2^1 =hy$ & $y_2^2 = 4 \times y_2^1 ** 2 + 0.3 $ &  \\ \hline
$X_3$ & $x_3^1 =hx$ & $x_3^2 = 2 \times x_3^1 ** 2 +\epsilon \sim \mathcal{N}(\epsilon; -0.3, 0.2)$ &  \\
$Y_3$ & $y_3^1 =hy$ & $y_3^2 = 0.5 \times y_3^1 ** 3 - 0.2 $ & 
\\
\midrule 
Data 4 & With distribution shift & With squared, cubed transformations or noises\\ 
$X_1$ & $x_1^1 =hx$ & $x_1^2 = x_1^1 +\epsilon\sim \mathcal{N}(\epsilon; 0, 0.3)$ &  \\
$Y_1$ & $y_1^1 =hy$ & $y_1^2 = y_1^1+\epsilon\sim \mathcal{N}(\epsilon; 0, 0.3)$ &  \\ \hline
$X_2$ & $x_2^1 =hx+\epsilon\sim \mathcal{N}(\epsilon; -0.1, 0.1)$ & $x_2^2 = 4 \times x_2^1 ** 3 + \epsilon \sim \mathcal{N}(\epsilon; 0.5, 0.3)$ &  \\
$Y_2$ & $y_2^1 =hy+\epsilon\sim \mathcal{N}(\epsilon; 0.2, 0.1)$  & $y_2^2 = 4 \times y_2^1 ** 2 + 0.3 $ &  \\ \hline
$X_3$ & $x_3^1 =hx+\epsilon\sim \mathcal{N}(\epsilon; 0.4, 0.2)$  & $x_3^2 = 2 \times x_3^1 ** 2 +\epsilon \sim \mathcal{N}(\epsilon; -0.3, 0.2)$ &  \\
$Y_3$ & $y_3^1 =hy+\epsilon\sim \mathcal{N}(\epsilon; -0.4, 0.2)$ & $y_3^2 = 0.5 \times y_3^1 ** 3 - 0.2 $ & \\
\bottomrule
\end{tabular}
\caption{Toy experiments: Synthetic data details for each experiment.}
\label{tab:syn}
\end{table*}
\begin{table*}[t]
    \centering
    \begin{tabular}{lccccccccc}
        \toprule
         Use ResNet-50 without SWAD &  $v2$  & $v3$ & $v1$  & lr mult & lr & dropout & WD & TR & CF \\ \hline
         TerraIncognita &  0.1 & 0.1 &  0.2 &  12.5 & - & - & - & - & - \\
         OfficeHome & 0.1 &  0.001 & 0.1 & 20.0 & 3e-5 & 0.1 & 1e-6 & - & - \\
         VLCS & 0.01 & 0.001 & 0.1 & 10.0 & 1e-5 & - & 1e-6 & 0.2 & 50 \\
         PACS & 0.01 & 0.01 & 0.01 & 25.0 & - & - & - & - & -   \\
         DomainNet & 0.1 & 0.1 & 0.1 & 7.5 & - & - & - & - & 500
         \\ \bottomrule 
         \\
    \end{tabular}

    \begin{tabular}{lccccccccc}
    \toprule
         Use ResNet-50 with SWAD & $v2$  & $v3$ & $v1$  & lr mult & CF \\ \hline
         TerraIncognita &  0.1 & 0.001 &  0.01 &  10.0 & - \\
         OfficeHome & 0.1 &  0.1 & 0.3 & 10.0 &  - \\
         VLCS & 0.01 & 0.001 & 0.1 & 10.0 &  50 \\
         PACS & 0.01 & 0.001 & 0.1 & 20.0 & -   \\ 
         DomainNet & 0.1 & 0.1 & 0.1 & 7.5 & 500 \\
    \bottomrule \\
    \end{tabular}

    \begin{tabular}{lccccccccc}
    \toprule
         Use RegNetY-16GF with and without SWAD & $v2$  & $v3$ & $v1$  & lr mult & CF \\ \hline
         TerraIncognita &  0.01 & 0.01 &  0.01 & 2.5 & - \\
         OfficeHome & 0.01 &  0.1 & 0.1 & 0.1 &  - \\
         VLCS & 0.01 &  0.01  & 0.1 & 2.0 &  50 \\
         PACS & 0.01 & 0.1 & 0.1 & 0.1 & -   \\ 
         DomainNet & 0.1 & 0.1 & 0.1 & 7.5 & 500 \\
    \bottomrule     
    \end{tabular}
  \caption{Classification experiments: Parameter settings of classification tasks. Notations: WD: weight decay; TR: tolerance ratio; CF: checkpoint freq. - denotes that for where the default settings are used.}
    \label{tab:main_para}
\end{table*}
\begin{table*}[t]
    \centering
     \begin{tabular}{lccccccccc}
    \toprule
         Ablation studies on OfficeHome& $v2$  & $v3$ & $v1$  & lr mult  & use \textbf{iAim1} & use \textbf{iReg2} \\ \hline
        {Base} (ERM) & 0.0 &  0.0 & 0.0 & 0.1 &  False & False\\
        {Base +iAim1} (DANN) & 0.0 &  0.0 & 0.1 & 0.1 &  True & False\\
        {Base + GAim1} (CDANN, CIDG)  & 0.0 &  0.0 & 0.1 & 0.1 &  False & False\\
        {Base  +iReg2} (CORAL+$\psi$) & 0.0 &  0.1 & 0.0 & 0.1 &  False & True\\
        {Base + GReg2} & 0.0 &  0.1 & 0.0 & 0.1 &  False & False\\
        {Base + GAim1 + GReg2} (MDA+$\psi$) & 0.0 &  0.1 & 0.1 & 0.1 &  False & False\\
        {Base + GReg1} (MIRO, SIMPLE) & 0.01 &  0.0 & 0.0 & 0.1 &  False & False\\
        {Base + GReg1 +iAim1} & 0.01 &  0.0 & 0.1 & 0.1 &  False & True\\
        {Base + GReg1 + GAim1} & 0.01 &  0.0 & 0.1 & 0.1 &  False & False\\
        {Base + GReg1 +iReg2} & 0.01 &  0.1 & 0.0 & 0.1 &  True & False\\
        {Base + GReg1 + GReg2} & 0.01 &  0.1 & 0.0 & 0.1 &  False & False\\
        {Base + GReg1 + GAim1 + GReg2 (Ours)} & 0.01 &  0.1 & 0.1 & 0.1 &  False & False\\
    \bottomrule   
    \end{tabular}
\caption{Ablation studies: Parameter settings of ablation studies. Notations: WD:  CF: checkpoint freq. - denotes that for where the default settings are used.}
\label{tab:ablation_para}
\end{table*}

\subsection{Regression experiments: Monocular depth estimation details.}
We explore the efficacy of $\psi$ with GMDG by using toy monocular depth estimation experiments with NYU Depth V2 dataset~\cite{silberman2012indoor}.

\textbf{Datasets.}
NYU Depth
V2 contains images with  $480 \times 640$  resolution with depth values ranging from 0 to 10 meters. We adopt the densely labeled pairs for training and testing.

\textbf{Multi-domain construction.} To construct multiple domains that fit the problem settings, we split the NYU Depth V2 dataset into four categories as four domains:
\begin{itemize}
    \item {School}: study room,
            study,
            student lounge,
            printer room,
            computer lab,
            classroom.
     \item {Office}: reception room,
            office kitchen,
            office,
            nyu office,
            conference room.
     \item {Home}: playroom,
            living room,
            laundry room,
            kitchen,
            indoor balcony,
            home storage,
            home office,
            foyer,
            dining room,
            dinette,
            bookstore,
            bedroom,
            bathroom,
            basement.
    \item {Commercial}:furniture store,
            exercise room,
            cafe.
\end{itemize}
After filtering data samples that are not able to be used, each domain has $95$, $110$, $1209$, and $35$ data pairs that can be used for training. 

\textbf{Parameter settings.} We follow all the hyperparameter settings in the VA-DepthNet and set $v_{A1} =0.001, v_{R1} = 0.001, v_{R2} = 0.0001$.  Note that the backbone is trained using VA-DepthNet but without the Variational Loss proposed by VA-DepthNet. 

\textbf{Experimental settings.} We use the final saved checkpoint for the leave-one-out cross-validation. 

\textbf{Metrics.} Please see metric details in VA-DepthNet~\cite{liu2023va}.

\subsection{Segmentation experimental details.}
We follow the experimental settings of RobustNet for segmentation experiments.

\textbf{Datasets.} There are two groups of datasets: Synthetic datasets and real-world datasets.
(1) Synthetic datasets: GTAV~\cite{richter2016gtav} is a large-scale dataset containing 24,966 driving-scene images generated from the Grand Theft Auto V game engine. SYNTHIA~\cite{ros2016synthia} which is composed of photo-realistic synthetic images has 9,400 samples with a resolution of 960×720.
(2) Real-world datasets: Cityscapes~\cite{cordts2016cityscapes} is a large-scale dataset containing high-resolution urban scene images. 
Providing 3,450 finely annotated images and 20,000 coarsely annotated images, it collects data from 50 different cities in primarily Germany. Only the finely annotated set is adopted for our training and validation. BDD-100K~\cite{yu2020bdd100k} is
another real-world dataset with a resolution of 1280×720. It provides diverse urban driving scene images from various locations in the US. We use the 7,000 training and 1,000 validation of the semantic segmentation task.
The images are collected from various locations in the US. 
Mapillary is also a real-world dataset that contains worldwide street-view scenes with 25,000 high-resolution images. 

\textbf{Parameter settings.} Specifically, we use all RobustNet's hyper-parameters and set $v_{A1} = 0.0001, v_{R1} = 0.0001$.

\subsection{Classification experimental details.} 


\textbf{Datasets.}
We use PACS (4 domains, 9,991
samples, $7$ classes)~\citep{li2017deeper}, VLCS ($4$ domains, $10,729$ samples, $5$ classes)~\citep{fang2013unbiased}, OfficeHome (4 domains, 15,588 samples, 65 classes)~\citep{venkateswara2017deep}, TerraIncognita (TerraInc, $4$ domains, $24,778$ samples, $10$ classes)~\citep{beery2018recognition}, and DomainNet (6 domains,
586,575 samples, 345 classes)~\citep{peng2019moment}.

\textbf{Parameter settings.}  We list the hyper-parameters in Table~\ref{tab:main_para} to reproduce our results.

\textbf{Metric.} We employ mean
Intersection over Union (mIoU) as the measurement for the segmentation task.

\subsection{Ablation studies experimental details.} 


\textbf{Parameter settings.} We run each experiment in three trials with seeds: $[0, 1, 2]$.  Full settings are reported in Table~\ref{tab:ablation_para}.
Especially,

\textbf{Experimental settings.} We use SWAD for all ablation studies to alleviate the effeteness of hyper-parameters. All ablation studies share the same hyper-parameters but add different combinations of terms. CORAL's~\citep{sun2016deep} objective focuses on minimizing the learned feature covariance discrepancy between source and target, requiring target data access and only regards second-order statistics. We adapt its approach to minimize feature covariances across seen domains for a fair comparison. 

\section{More results}
\label{app:res}

\begin{figure}[t]
    \centering
    \includegraphics[width=\linewidth]{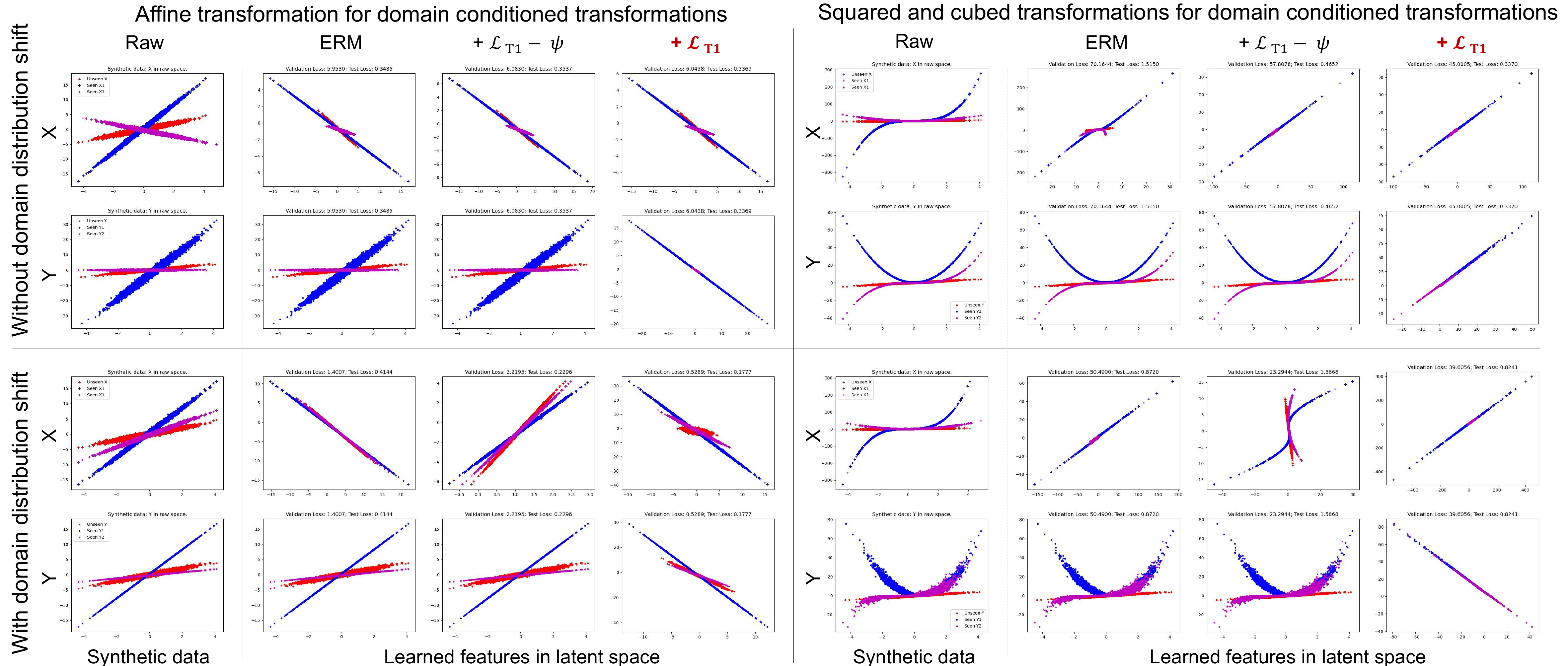}
    \caption{Toy experiments: Visualization of learned latent representations of different methods. Each color represents a domain.}
    \label{fig:toy_res}
\end{figure}

\textbf{Visualization of toy experiments:} Please see the visualization of toy experiments in \Figref{fig:toy_res}.

\textbf{Regression results: Monocular
depth estimation.}

\begin{figure*}[t]
    \centering
    \includegraphics[width=\linewidth]{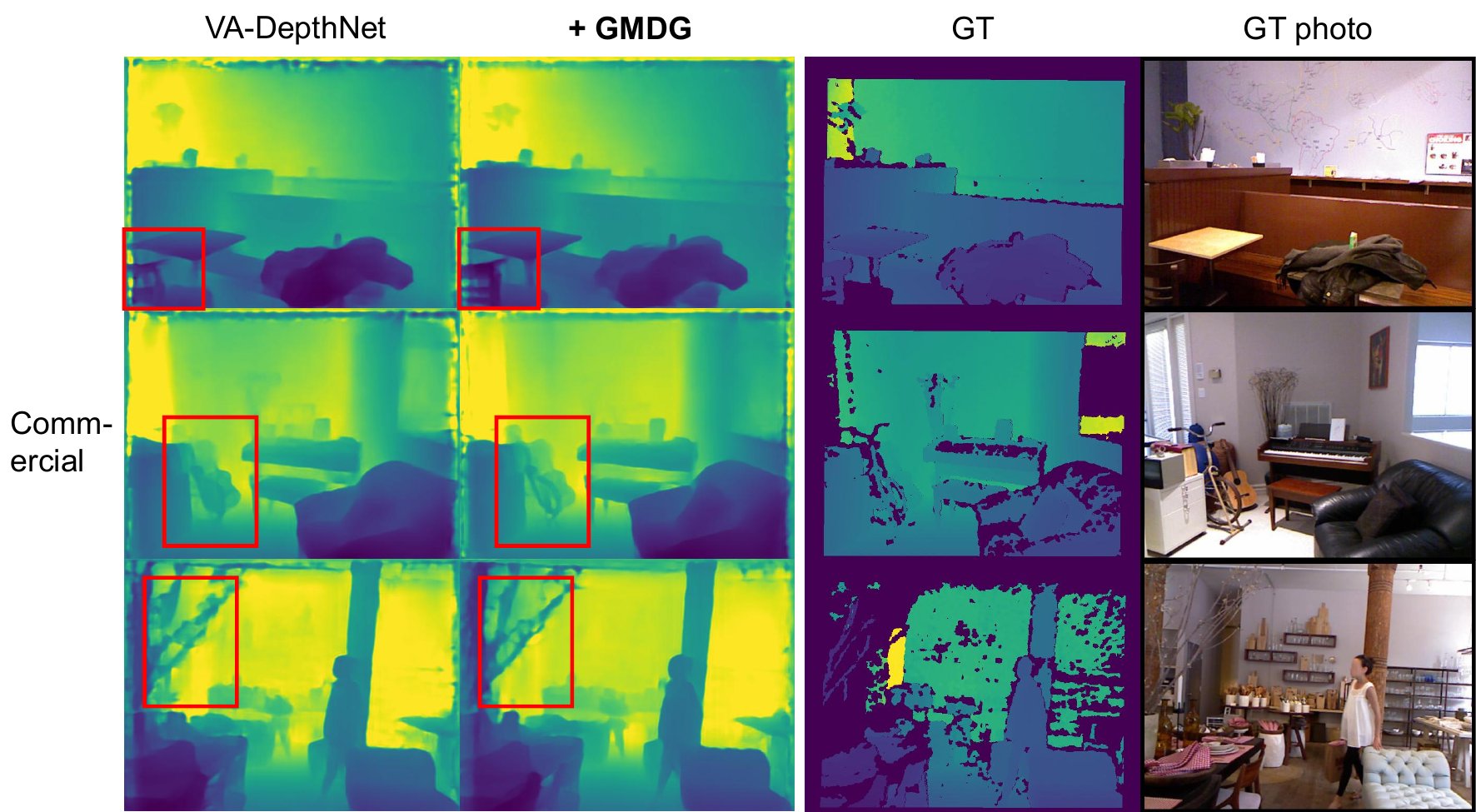}
    ~\\~\\
    \includegraphics[width=\linewidth]{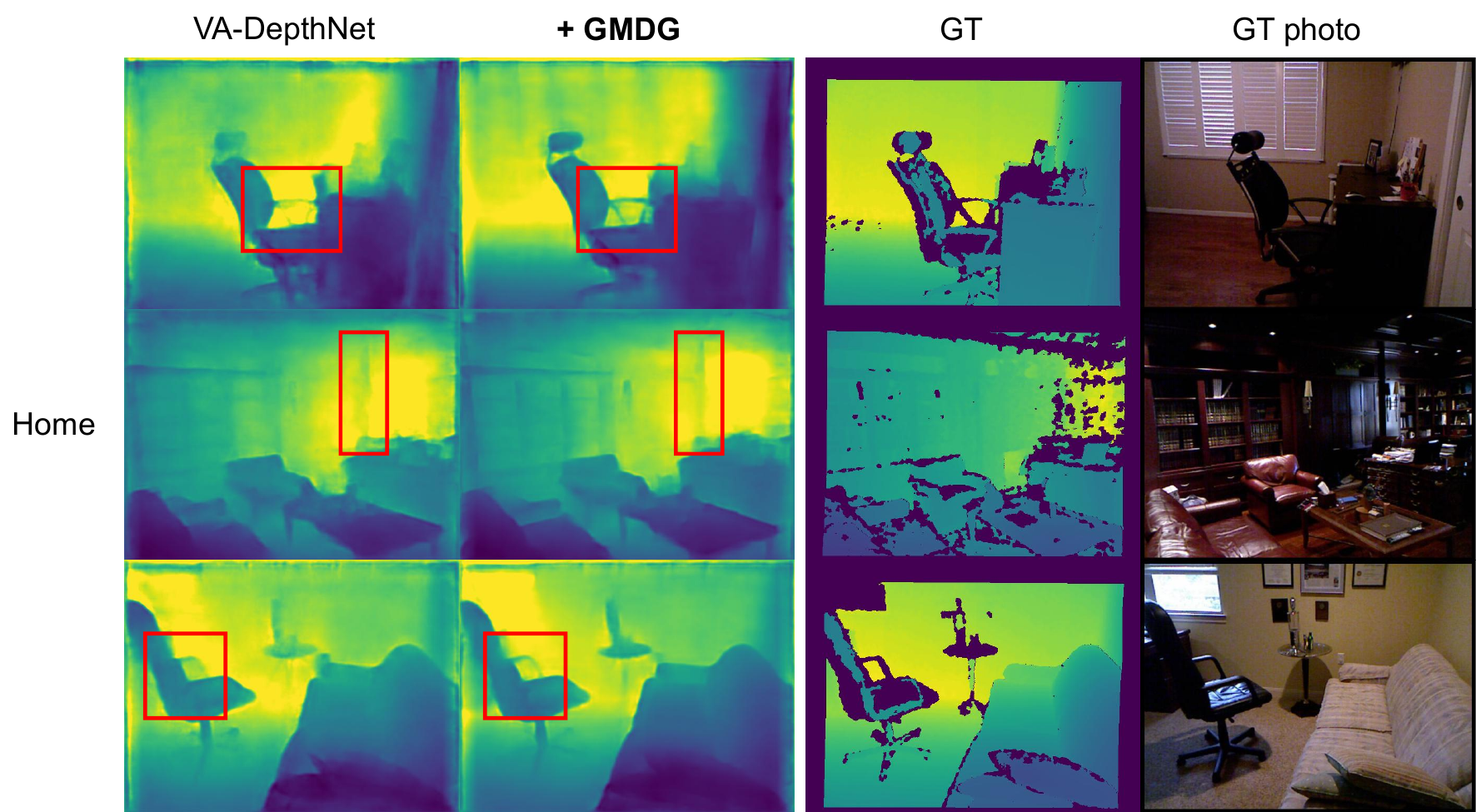}
    \caption{Regression results: Monocular depth estimation results between VA-Depth and our GMDG on samples from unseen domains. It can be seen that better generalization across domains is obtained with GMDG.}
    \label{fig:reg_res}
\end{figure*}

\begin{figure*}[t]
    \centering
    \includegraphics[width=\linewidth]{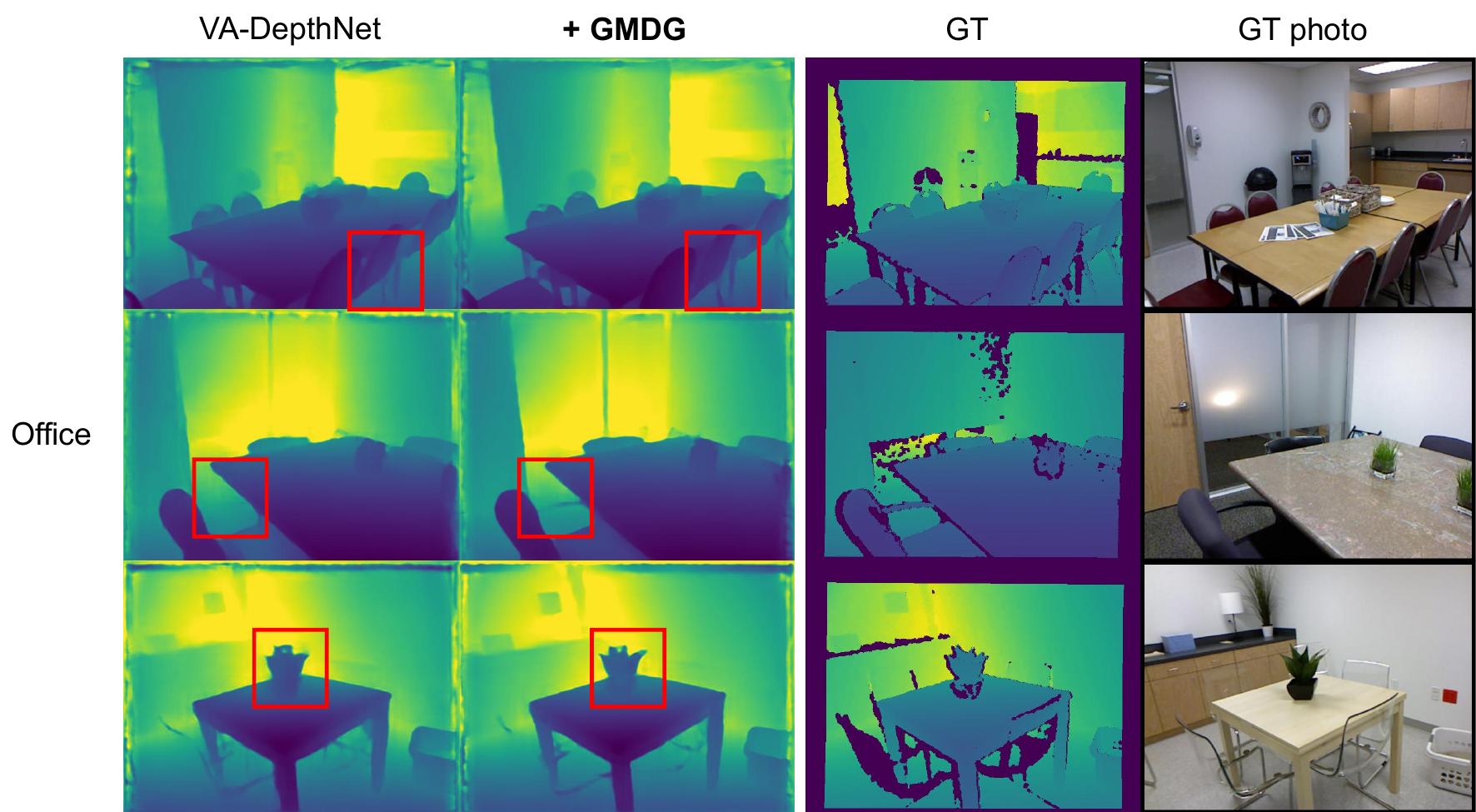}
    ~\\~\\
    \includegraphics[width=\linewidth]{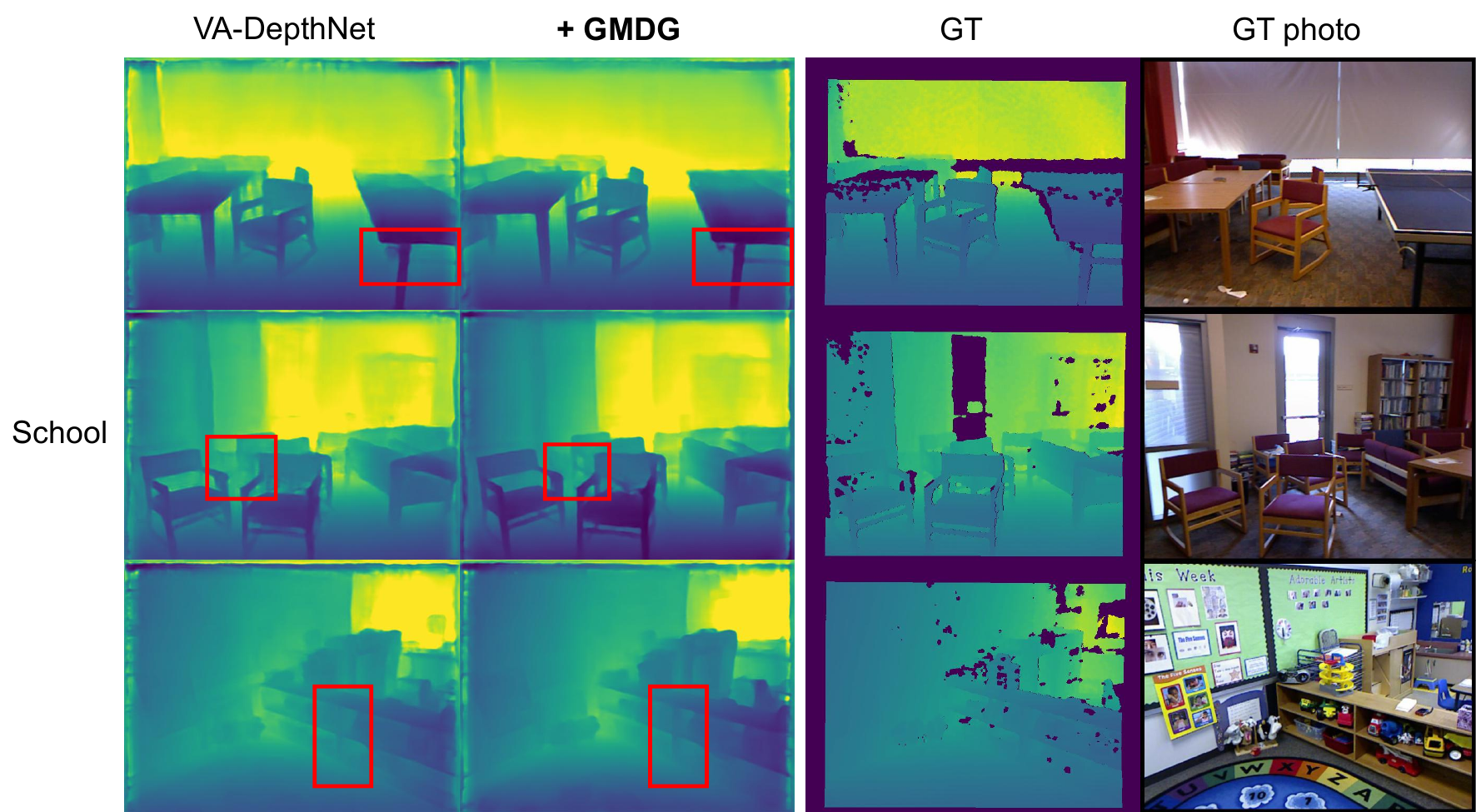}
    \caption{Regression results: Monocular depth estimation results between VA-Depth and our GMDG on samples from unseen domains. It can be seen that better generalization across domains is obtained with GMDG, continues.}
    \label{fig:reg_res2}
\end{figure*}

The regression results for each unseen domain of monocular
depth estimation visualization is displayed in Figure~\ref{fig:reg_res},~\ref{fig:reg_res2}.

\begin{figure*}[t]
    \centering
    \includegraphics[width=\linewidth]{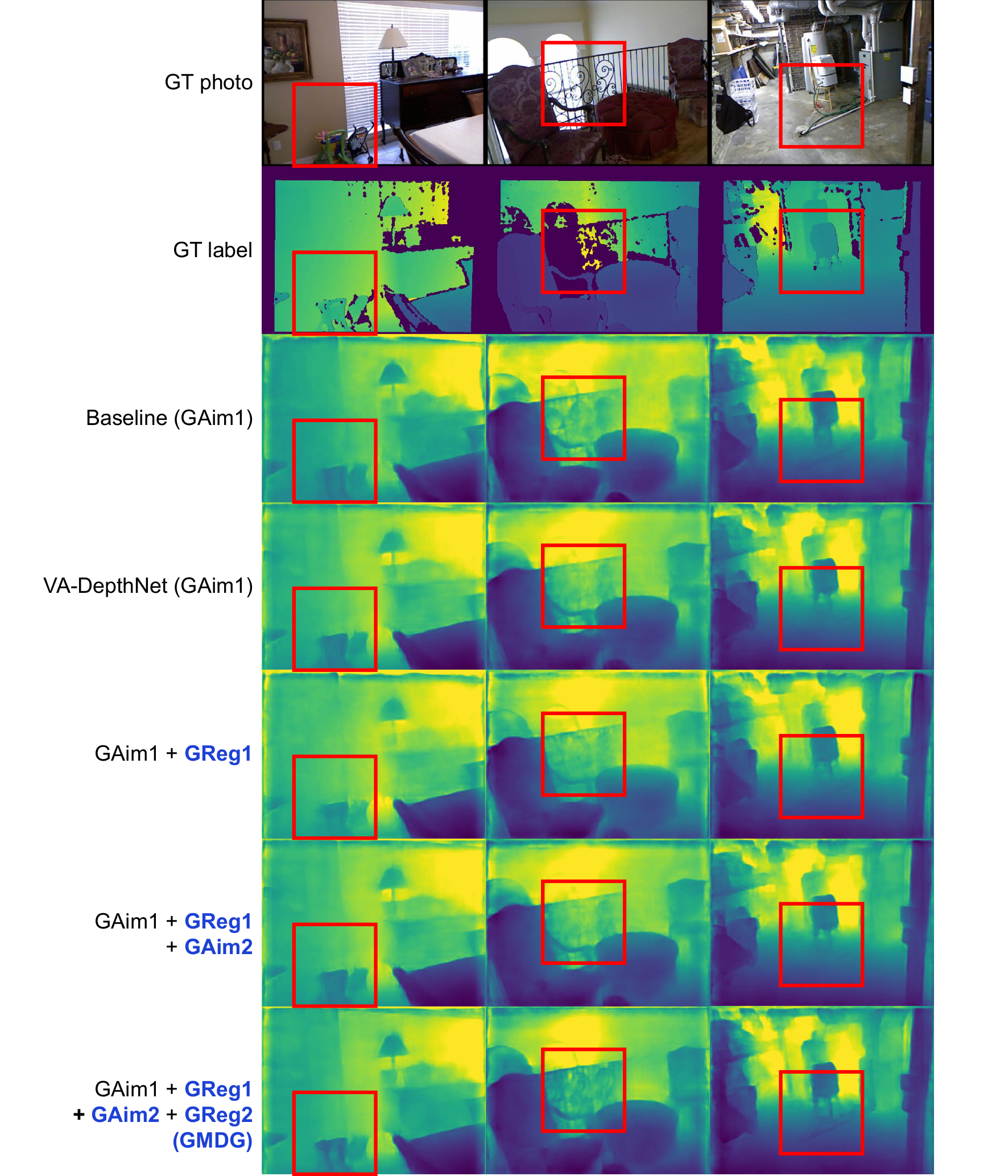}
    \caption{Regression results: Monocular depth estimation results for unseen domains of models trained with different objective settings. It can be seen that with the whole GMDG, the model performs the best generalization for all unseen domain settings.  }
    \label{fig:reg_abl_res1}
\end{figure*}

\begin{figure*}[t]
    \centering
    \includegraphics[width=\linewidth]{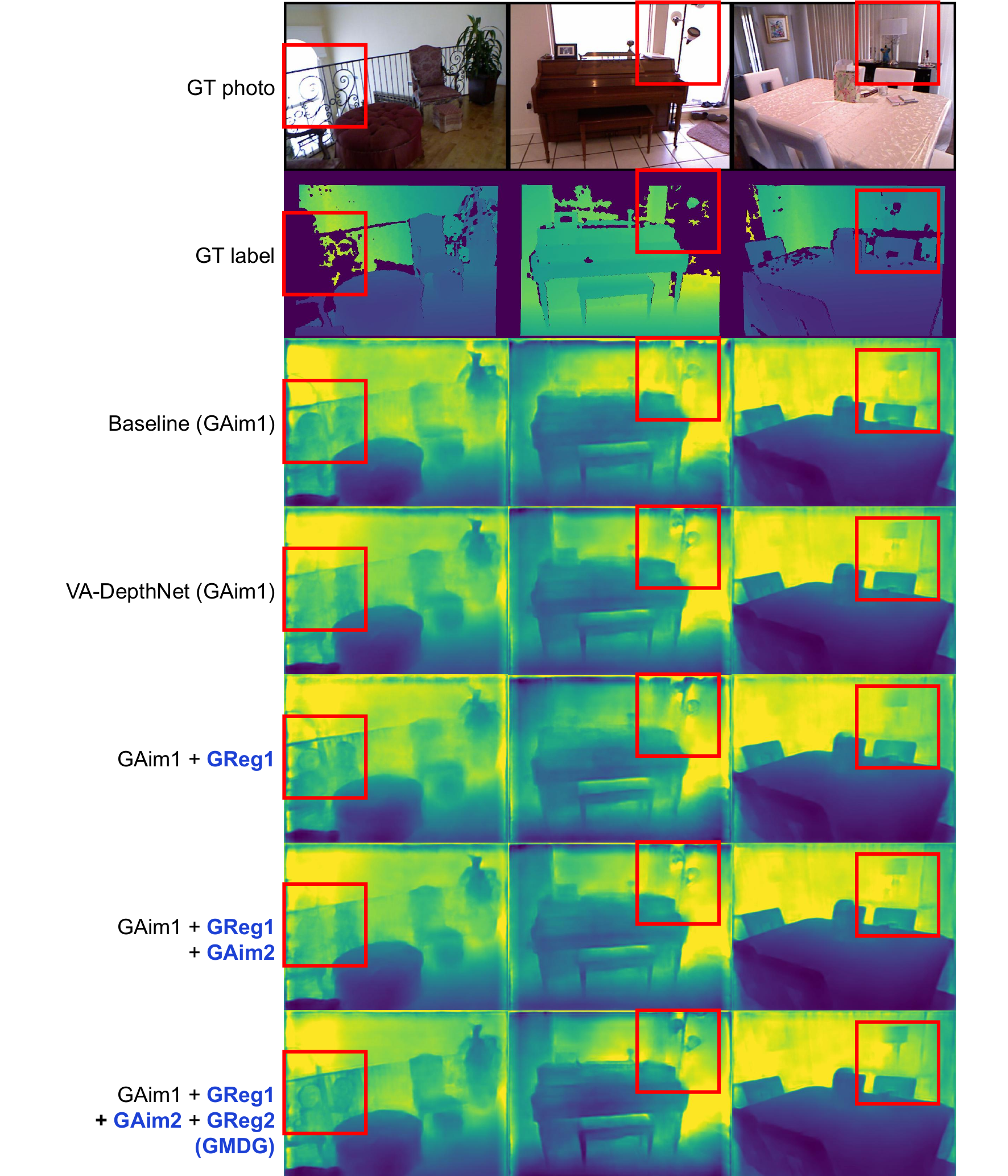}
    \caption{Regression results: Monocular depth estimation results for unseen domains of models trained with different objective settings, continues. It can be seen that with the whole GMDG, the model performs the best generalization for all unseen domain settings.}
    \label{fig:reg_abl_res2}
\end{figure*}

The Visualization of regression results for unseen domains of models trained with different objective settings are exhibited in  Figure~\ref{fig:reg_abl_res1},~\ref{fig:reg_abl_res2}.

\begin{figure*}
    \centering
    \includegraphics[width=0.9\linewidth]{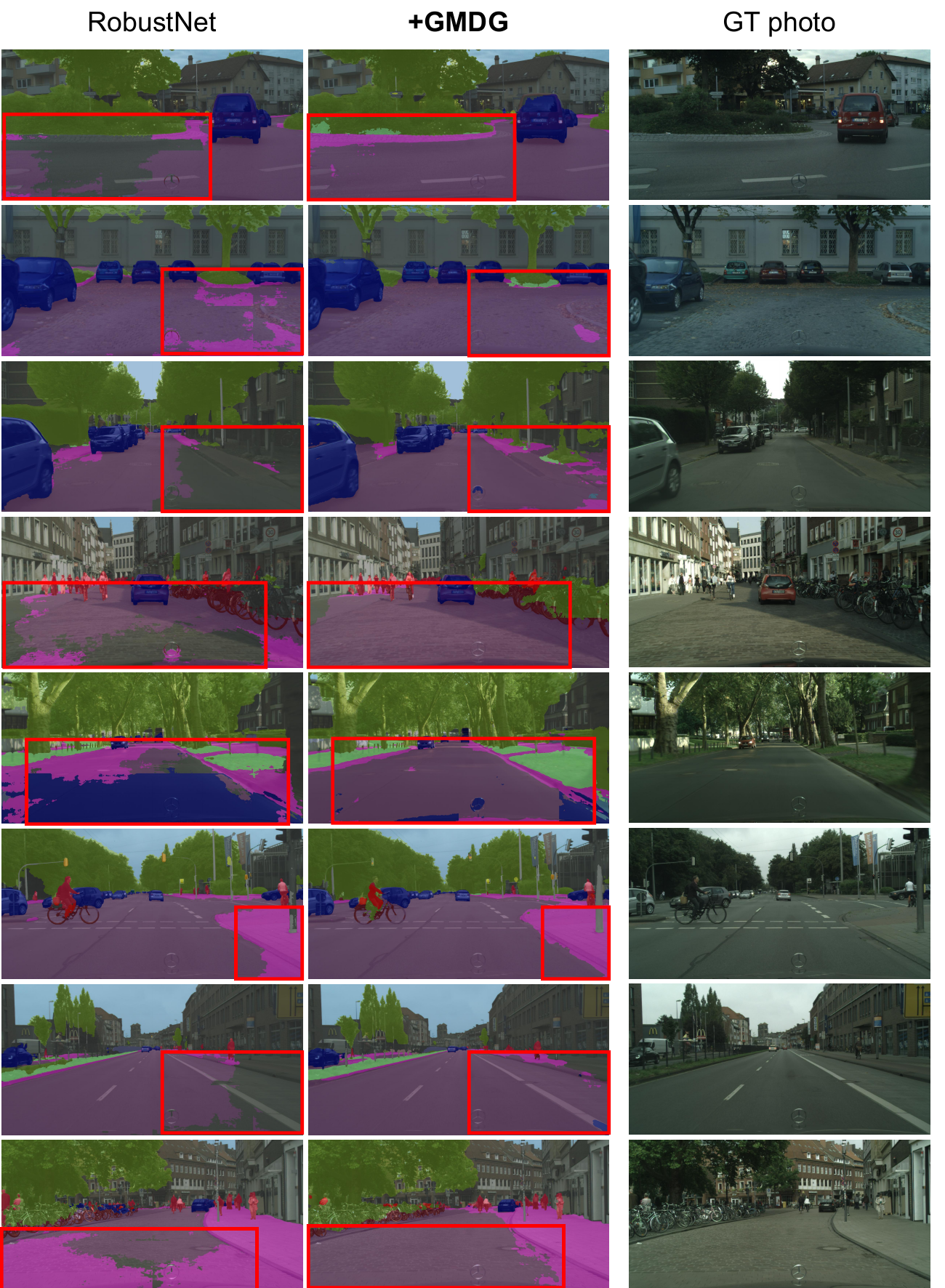}
    \caption{Segmentation results: Visualizations between RobostNet and our GMDG on samples from unseen domains. It can be seen that better generalization is obtained with GMDG.}
    \label{fig:seg_res}
\end{figure*}

\textbf{Segmentation results.}
The segmentation results for unseen samples are displayed in \Figref{fig:seg_res}.

\textbf{Classification results.}
We show the results of each category for the classification experiments as Table~\ref{tab:res_detial1},~\ref{tab:res_detial2},~\ref{tab:res_detial3},~\ref{tab:res_detial4},~\ref{tab:res_detial5}.

\subsection{Other findings and Analysis}
\label{app:other_findings}
\textbf{What makes a better $\mathcal{O}$.}
    As demonstrated in Eq.~\ref{eq:PUB}, $\mathcal{O}$ plays a crucial role in PUB by anchoring a space where the relationship between $\rmX$ and $\rmY$ is preserved. Ideally, having one $\mathcal{O}$ that provides general representations for all seen and unseen domains leads to the best results, one finding supported by MIRO and SIMPLE. However, even though SIMPLE++ combines 283 pre-trained models, achieving the `perfect' $\mathcal{O}$ remains unattained. Therefore, this paper primarily discusses how our proposed objectives can improve the model performance when a fixed $\mathcal{O}$ is provided.

\textbf{Comparison with MDA: Minimizing domain gap compared to the decision gap.} 
MDA~\citep{hu2020domain}, guided by the hypothesis ``guaranteed generalization only when the decision gap exceeds the domain gap", aims to minimize the ratio between the domain gap and the decision gap. This approach facilitates learning $\mathcal{D}$-independent conditional features, enhancing class separability across domains.
As Table~\ref{tab:summary} illustrates, MDA's \textbf{Reg2} objective can also be interpreted as suppressing invalid causality, aligning with our approach.
However, MDA's implementation requires manual selection of $\phi(\rmX)$ from the same $\rmY$ without using $\psi$ and \textbf{GReg2}. Our method further relaxes MDA's assumption, extending the application of the objective and making it also applicable to tasks besides classification, such as segmentation.  

\textbf{Cutting off causality form $\phi(\rmX) \to \psi(\rmY)$ may lead to collapse of the model.} We have tried to reversely suppress the causality form $\phi(\rmX) \to \psi(\rmY)$ instead of causality form $\psi(\rmY) \to \phi(\rmX)$ for monocular depth estimation in VA-DepthNet and it causes collapse.

\begin{figure*}[ht]
  \centering
    \includegraphics[width=0.8\linewidth]{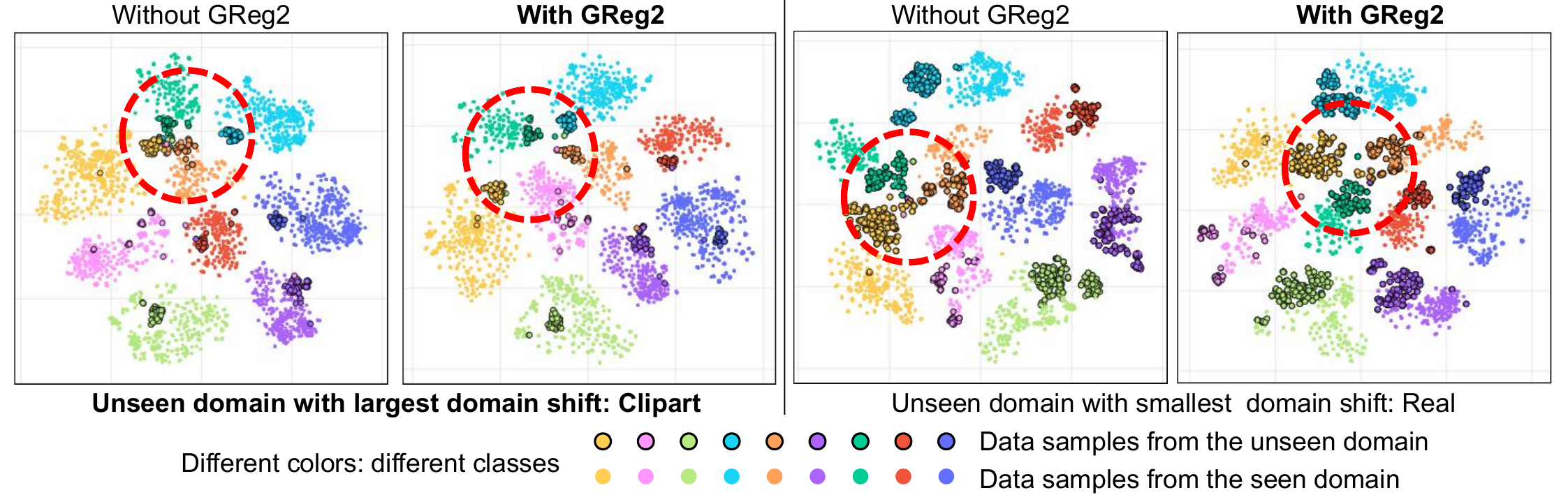}
   \caption{T-SNE map of latent features from classification models that were trained without and with \textbf{GReg2}.}
   \label{fig:GReg2}
\end{figure*}

\textbf{{Suppressing invalid causality:}}
\textbf{{Why this design:}}
Our GMDG introduces a mapping $\psi$ for $\rmY$ to relax the static assumption, corroborating more general and practical scenarios. 
Our empirical findings, as shown in
\cref{fig:GReg2}, 
reveal that
introducing 
$\psi(\rmY)$ without any constraints may not guarantee a clear decision margin for classification. 
Upon examining our objective in Eq.8 in the main manuscript, we hypothesize that 
the effect might result from
`$\psi(\rmY)$ causing $\phi(\rmX)$', which we term `invalid causality'. 
Thus, we designed a term to suppress such invalid causality. 
This term is derived from the prediction perspective wherein $\rmY$ should be only predicted from $\phi(\rmX)$; hence, $\phi(\rmX)$ 
should not be caused by $\psi(\rmY)$. 
Consider the scenario wherein $\rmY$ and $\psi(\rmY)$ are absent during prediction - the hypothesized causality from $\psi(\rmY)$ to $\phi(\rmX)$ would disrupt the causal chain, resulting in an `incomplete' representation of $\phi(\rmX)$ then prediction degradation. 
Hence, it is critical to suppress  $\psi(\rmY) \!\!\!\to\!\!\! \phi(\rmX)$ that may occur during joint training. Notably, the suppression is not symmetric and promotes $\phi(\rmX) \!\!\to\!\! \rmY$.
\textbf{{Intuition:}} 
Intuitively, \textbf{GReg2} further `erases' the redundant information in $\phi(\rmX)$ that may be caused by $\phi(\rmY)$, which aims to refine the latent space, yielding 
better invariant latent features for predicting $\rmY$ (i.e., larger decision margin for the unseen domain as highlighted in \cref{fig:GReg2}).

\textbf{{GMDG's efficiency:}}
We have analyzed the efficiency of GMDG in \cref{tab:flop_param}. Though theoretically superior in generality, GMGD may increase computational costs during training due to additional loss functions and VAE encoders. However, these auxiliary components are discarded in the inference stage, ensuring their efficiency remains unaffected. 
Our confidence in the model's amenability to efficiency enhancement through careful 
design is high, and such pursuits remain a promising avenue for future work.
 



\begin{table*}[t]
\centering
\begin{tabular}{l|ll|ll|ll|l}
\hline
                                 & \multicolumn{2}{l|}{Classification}          & \multicolumn{2}{l|}{Depth estimation}        & \multicolumn{2}{l|}{Segmentation}            &           \\
                                 & Training & \cellcolor[HTML]{D5D5D5}Inference & Training & \cellcolor[HTML]{D5D5D5}Inference & Training & \cellcolor[HTML]{D5D5D5}Inference &           \\ \hline
                                 & 99.16    & \cellcolor[HTML]{D5D5D5}49.58     & 584.44   & \cellcolor[HTML]{D5D5D5}573.78    & 209.97   & \cellcolor[HTML]{D5D5D5}167.34    & Baseline  \\
\multirow{-2}{*}{FLOPs (G)}      & 124.00   & \cellcolor[HTML]{D5D5D5}49.58     & 1543.90  & \cellcolor[HTML]{D5D5D5}573.78    & 449.83   & \cellcolor[HTML]{D5D5D5}167.34    & With GMDG \\ \hline
                                 & 2.94     & \cellcolor[HTML]{D5D5D5}1.47      & 64.42    & \cellcolor[HTML]{D5D5D5}64.27     & 23.13    & \cellcolor[HTML]{D5D5D5}22.54     & Baseline  \\
\multirow{-2}{*}{Parameters (M)} & 4.93     & \cellcolor[HTML]{D5D5D5}1.47      & 123.76   & \cellcolor[HTML]{D5D5D5}64.27     & 82.48    & \cellcolor[HTML]{D5D5D5}22.54     & With GMDG \\ \hline
\end{tabular}%
\vspace{-0.2cm}
\caption{FLOPs and parameters of baselines without and with GMDG during training and inference.}
\label{tab:flop_param}
\end{table*}

\textbf{{Applicability, constraint, and limitations:}}
GMDG is specifically designed for mDG with accessible $P(\rmY)$, in which the model is trained on multiple seen domains and tested on one unseen domain. This task essentially requires learning the invariance across multi-domains for the prediction.  
When used in single-domain generalization or cases involving novel classes in the unseen domain, GMDG  may not be directly applicable, thus requiring further investigations. 
Meanwhile, as a general objective,  our novel GMDG involves 
additional modules/losses that may incur extra computational costs during training. We have discussed these aspects in our main paper, and we would like to leave them as future work.


\begin{table*}[t]
\small
\centering
\begin{tabular}{l|cccc|c}
\toprule
 \textbf{TerraIncognita} & \multicolumn{1}{c}{\textbf{Location 100}} & \multicolumn{1}{c}{\textbf{Location 38}} & \multicolumn{1}{c}{\textbf{Location 43}} & \multicolumn{1}{c|}{\textbf{Location 46}} & \multicolumn{1}{c}{\textbf{Avg.}} \\ \midrule
ERM~\cite{gulrajani2020search} & 54.3 & 42.5 & 55.6 & {38.8} & 47.8\\
MIRO~\cite{cha2022miro} (use ResNet-50) & - & -&- &- &{50.4} \\
\rowcolor{mygray}\textbf{GMDG} (use ResNet-50) & 
 60.9±5.34	& 47.3±3.42	& 55.2±1.06 &	41.0±2.93	& 51.1±0.91  \\ 		
\midrule
ERM + SWAD~\cite{cha2021swad} & 55.4 & 44.9 & 59.7 & 39.9 & 50.0 \\
DIWA~\cite{rame2022diverse} & 57.2&  {50.1} & {60.3} & 39.8 & 51.9 \\
MIRO~\cite{cha2022miro} + SWAD~\cite{cha2021swad}   (use ResNet-50) & - & -&- &- &52.9 \\
\rowcolor{mygray}\textbf{GMDG} + SWAD (use ResNet-50)& {61.2±1.4} & 48.4±1.6 & 60.0±0.4 & {42.5±1.1} & {53.0±0.7} \\ \midrule
MIRO~\cite{cha2022miro}  (use RegNetY-16GF) & - & -&- &- &58.9 \\
\rowcolor{mygray}\textbf{GMDG} (use RegNetY-16GF) & 73.3±3.3 & 54.7±1.4 & 67.1±0.3 & 48.6±6.5 & 60.7±1.8 \\ \midrule
MIRO~\cite{cha2022miro} + SWAD~\cite{cha2021swad}  (use RegNetY-16GF) & - & -&- &- &64.3 \\
\rowcolor{mygray}\textbf{GMDG} + SWAD (use RegNetY-16GF)& \textbf{74.3±1.5} & \textbf{59.2±1.2} & \textbf{70.6±1.1} & \textbf{56.0±0.8} & \textbf{65.0±0.2} \\ 
\bottomrule
\end{tabular}
\caption{Classification experiments on TerraIncognita: More results of full GMDG for each category.}
\label{tab:res_detial1}
\end{table*}

\begin{table*}[t]
\small
\centering
\begin{tabular}{l|cccc|c}
\toprule
 \textbf{OfficeHome} & \multicolumn{1}{c}{\textbf{art}} & \multicolumn{1}{c}{\textbf{clipart}} & \multicolumn{1}{c}{\textbf{product}} & \multicolumn{1}{c}{\textbf{real}} & \multicolumn{1}{c}{\textbf{Avg.}}  \\ \midrule
ERM~\cite{gulrajani2020search} & 63.1 & 51.9  & 77.2&  78.1& 67.6\\
MIRO~\cite{cha2022miro} (use ResNet-50) & - & -&- &- &70.5±0.4 \\
\rowcolor{mygray}\textbf{GMDG} (use ResNet-50) & 68.9±0.3 & 56.2±1.7 & 79.9±0.6 & 82.0±0.4 & 70.7±0.2  \\
\midrule
ERM + SWAD~\cite{cha2021swad} &66.1 &57.7 &78.4 &80.2& 70.6 \\
DIWA~\cite{rame2022diverse} & 69.2 & 59 & 81.7& 82.2 &72.8 \\
MIRO~\cite{cha2022miro} + SWAD~\cite{cha2021swad}   (use ResNet-50) & - & -&- &- &72.4±0.1 \\
\rowcolor{mygray}\textbf{GMDG} + SWAD (use ResNet-50)& 68.9±0.6 & 58.2±0.6 & 80.4±0.3 & 82.6±0.4 & 72.5±0.2 \\ \midrule
MIRO~\cite{cha2022miro}  (use RegNetY-16GF) & - & -&- &- & 80.4±0.2 \\
\rowcolor{mygray}\textbf{GMDG} (use RegNetY-16GF) & 79.7±1.6 & 67.7±1.8 & 87.8±0.8 & 87.9±0.7 & 80.8±0.6 \\ \midrule
MIRO~\cite{cha2022miro} + SWAD~\cite{cha2021swad}  (use RegNetY-16GF) & - & -&- &- &83.3±0.1 \\
\rowcolor{mygray}\textbf{GMDG} + SWAD (use RegNetY-16GF)& \textbf{84.1±0.2} & \textbf{74.3±0.9} & \textbf{89.9±0.4} & \textbf{90.6±0.1} & \textbf{84.7±0.2} \\ 
\bottomrule
\end{tabular}
\caption{Classification experiments on OfficeHome: More results of full GMDG for each category.}
\label{tab:res_detial2}
\end{table*}

\begin{table*}[t]
\small
\centering
\begin{tabular}{l|cccc|c}
\toprule
\textbf{VLCS} & \multicolumn{1}{c}{\textbf{caltech101}} & \multicolumn{1}{c}{\textbf{labelme}} & \multicolumn{1}{c}{\textbf{sun09}} & \multicolumn{1}{c|}{\textbf{voc2007}} & \multicolumn{1}{c}{\textbf{Avg.}}\\ \midrule
ERM~\cite{gulrajani2020search}& 97.7 &64.3& 73.4 &74.6 &77.3\\
MIRO~\cite{cha2022miro} (use ResNet-50) & - & -&- &- & 79.0±0.0 \\
\rowcolor{mygray}\textbf{GMDG} (use ResNet-50)& 98.3±0.4 & 65.9±1 & 73.4±0.8 & 79.3±1.3 & 79.2±0.3\\
\midrule
ERM + SWAD~\cite{cha2021swad}&98.8& 63.3& 75.3 &79.2& 79.1\\
DIWA~\cite{rame2022diverse}&98.9& 62.4 &73.9 &78.9 &78.6 \\
MIRO~\cite{cha2022miro} + SWAD~\cite{cha2021swad}(use ResNet-50) & - & -&- &- &  79.6±0.2\\
\rowcolor{mygray}\textbf{GMDG} + SWAD (use ResNet-50) & 98.9±0.4 & 63.6±0.2 & 76.4±0.5 & 79.5±0.6 & 79.6±0.1 \\ \midrule
MIRO~\cite{cha2022miro}  (use RegNetY-16GF) & - & -&- &- & 79.9±0.6\\
\rowcolor{mygray}\textbf{GMDG} (use RegNetY-16GF) & 97.9±1.3 & \textbf{66.8±2.1} & \textbf{80.8±1} & 83.9±1.8 & \textbf{82.4±0.6} 
\\ \midrule
MIRO~\cite{cha2022miro} + SWAD~\cite{cha2021swad}  (use RegNetY-16GF) & - & -&- &- & 81.7±0.1\\
\rowcolor{mygray}\textbf{GMDG} + SWAD (use RegNetY-16GF)  & \textbf{98.4±0.1} & 65.5±1.4 & 79.9±0.4 & \textbf{84.9±0.9} & 82.2±0.3\\ 
\bottomrule
\end{tabular}
\caption{Classification experiments on VLCS: More results of full GMDG for each category.}
\label{tab:res_detial3}
\end{table*}

\begin{table*}[t]
\small
\centering
\begin{tabular}{l|cccc|c}
\toprule
 \textbf{PACS} & \multicolumn{1}{c}{\textbf{art\_painting}} & \multicolumn{1}{c}{\textbf{cartoon}} & \multicolumn{1}{c}{\textbf{photo}} & \multicolumn{1}{c|}{\textbf{sketch}} & \multicolumn{1}{c}{\textbf{Avg.}}  \\ \midrule
ERM~\cite{gulrajani2020search}  &84.7 & 80.8 & 97.2 & 79.3  &84.2\\
MIRO~\cite{cha2022miro} (use ResNet-50)& - & -&- &- & 85.4±0.4 \\
\rowcolor{mygray}\textbf{GMDG} (use ResNet-50) & 84.7±1.0 & 81.7±2.4 & 97.5±0.4 & 80.5±1.8 & 85.6±0.3\\
\midrule
ERM + SWAD~\cite{cha2021swad} &89.3 &83.4& 97.3& 82.5 &88.1\\
DIWA~\cite{rame2022diverse} &90.6 &83.4 &98.2 &83.8 &89.0\\
MIRO~\cite{cha2022miro} + SWAD~\cite{cha2021swad}   (use ResNet-50) & - & -&- &- & 88.4±0.1 \\
\rowcolor{mygray}\textbf{GMDG} + SWAD (use ResNet-50) & 90.1±0.6 & 83.9±0.2 & 97.6±0.5 & 82.3±0.7 & 88.4±0.1 \\ \midrule
MIRO~\cite{cha2022miro}  (use RegNetY-16GF)& - & -&- &- & 97.4±0.2\\
\rowcolor{mygray}\textbf{GMDG} (use RegNetY-16GF) & 97.5±1.0 & 97.0±0.2 & 99.4±0.2 & 95.2±0.4 & 97.3±0.1
\\ \midrule
MIRO~\cite{cha2022miro} + SWAD~\cite{cha2021swad}  (use RegNetY-16GF) & - & -&- &- &96.8±0.2 \\
\rowcolor{mygray}\textbf{GMDG} + SWAD (use RegNetY-16GF) & \textbf{98.3±0.3} & \textbf{98.0±0.1} & \textbf{99.5±0.3} & \textbf{95.3±0.8} & \textbf{97.9±0.0}\\ 
\bottomrule
\end{tabular}
\caption{Classification experiments on PACS: More results of full GMDG for each category.}
\label{tab:res_detial4}
\end{table*}

\begin{table*}[t]
\small
\centering
\begin{tabular}{l|cccccc|c}
\toprule
 \textbf{DomainNet} &  \multicolumn{1}{c}{\textbf{clipart}} & \multicolumn{1}{c}{\textbf{info}} & \multicolumn{1}{c}{\textbf{painting}} & \multicolumn{1}{c}{\textbf{quickdraw}} & \multicolumn{1}{c}{\textbf{real}} & \multicolumn{1}{c|}{\textbf{sketch}} & \multicolumn{1}{c}{\textbf{Avg.}} \\ \midrule
ERM~\cite{gulrajani2020search}&50.1 &63.0& 21.2 &63.7& 13.9& 52.9 &44.0 \\
MIRO~\cite{cha2022miro} (use ResNet-50)& - & -&- &- & - & -& 44.3±0.2\\
\rowcolor{mygray}\textbf{GMDG} (use ResNet-50) & 63.4±0.3 & 22.4±0.4 & 51.4±0.4 & 13.4±0.8 & 64.4±0.3 & 52.4±0.4 & 44.6±0.1 \\
\midrule
ERM + SWAD~\cite{cha2021swad} &53.5 &66.0& 22.4 &65.8& 16.1 &55.5 &46.5\\
DIWA~\cite{rame2022diverse} &55.4 &66.2& 23.3 &68.7& 16.5 &56.0 &47.7\\
MIRO~\cite{cha2022miro} + SWAD~\cite{cha2021swad}   (use ResNet-50)   & - & -&- &- & - & -&47.0±0.0\\
\rowcolor{mygray}\textbf{GMDG} + SWAD (use ResNet-50)  & 66.4±0.3 & 23.8±0.1 & 54.5±0.3 & 15.8±0.1 & 67.5±0.1 & 55.8±0.0 & 47.3±0.1\\ \midrule
MIRO~\cite{cha2022miro}  (use RegNetY-16GF) & - & -&- &- & - & -& 53.8±0.1\\
\rowcolor{mygray}\textbf{GMDG} (use RegNetY-16GF) & 74.0±0.3 & 39.5±1.5 & 61.5±0.3 & 16.3±1.2 & 73.9±1.5 & 62.8±2.4 & 54.6±0.1 
\\ \midrule
MIRO~\cite{cha2022miro} + SWAD~\cite{cha2021swad}  (use RegNetY-16GF) & - & -&- &- & - & -& 60.7±0.0 \\
\rowcolor{mygray}\textbf{GMDG} + SWAD (use RegNetY-16GF)  & \textbf{79.0±0.0} & \textbf{46.9±0.4} & \textbf{69.9±0.4} & \textbf{20.7±0.6} & \textbf{81.1±0.3} & \textbf{70.3±0.4} & \textbf{61.3±0.2}  \\ 
\bottomrule
\end{tabular}
\caption{Classification experiments on DomainNet: More results of full GMDG for each category.}
\label{tab:res_detial5}
\end{table*}

\end{document}